\documentclass{article}

\usepackage{PRIMEarxiv}

\usepackage[utf8]{inputenc} 
\usepackage[T1]{fontenc}    
\usepackage{hyperref}       
\usepackage{url}            
\usepackage{booktabs}       
\usepackage{amsfonts}       
\usepackage{nicefrac}       
\usepackage{microtype}      
\usepackage{lipsum}
\usepackage{fancyhdr}       
\usepackage{graphicx}       
\graphicspath{{media/}}     
\usepackage{wrapfig}

\usepackage{lscape}
\usepackage{longtable}

\usepackage{xltabular}
\usepackage{caption}
\usepackage{tabularray}
\usepackage{tablefootnote}
\usepackage{enumerate}
\usepackage{array}
\usepackage{orcidlink}
\usepackage{hyperref}
\title{A Comprehensive Survey on Rare Event Prediction
}

\author{
  Chathurangi Shyalika\orcidlink{0000-0002-5320-5566}\\
  AI Institute \\
  University of South Carolina \\
  USA\\
  \texttt{jayakodc@email.sc.edu} \\
   \And
   Ruwan Wickramarachchi\orcidlink{0000-0001-5810-1849}\\
  AI Institute \\
  University of South Carolina \\
  USA\\
  \texttt{ruwan@email.sc.edu} \\
   \And
  Amit Sheth\orcidlink{0000-0002-0021-5293}\\
  AI Institute \\
  University of South Carolina \\
  USA\\
  \texttt{amit@sc.edu} \\
}

\begin{document}
\maketitle

\begin{abstract}
Rare event prediction involves identifying and forecasting events with a low probability using machine learning (ML) and data analysis. Due to the imbalanced data distributions, where the frequency of common events vastly outweighs that of rare events, it requires using specialized methods within each step of the ML pipeline, i.e., from data processing to algorithms to evaluation protocols. Predicting the occurrences of rare events is important for real-world applications, such as Industry 4.0, and is an active research area in statistical and ML. This paper comprehensively reviews the current approaches for rare event prediction along four dimensions: rare event data, data processing, algorithmic approaches, and evaluation approaches. Specifically, we consider 73 datasets from different modalities (i.e., numerical, image, text, and audio), four major categories of data processing, five major algorithmic groupings, and two broader evaluation approaches. This paper aims to identify gaps in the current literature and highlight the challenges of predicting rare events. It also suggests potential research directions, which can help guide practitioners and researchers.
\end{abstract}


\section{Introduction}
Events are incidents that are associated with specific locations (spatial), time periods (temporal), and contexts (semantics). Rare events are a subset of events that stand out due to their infrequency. The degree of infrequency of rare events is typically influenced by the specific field of application  \cite{harrison2016rare, glasserman1999multilevel}. Rare event learning is considered an NP-hard problem \cite{weiss1998learning}, as it requires analyzing a large amount of data to identify rare events, which can be computationally intensive and time-consuming, especially in high-dimensional spaces. The size and complexity of rare event data lead to the challenge of handling this problem, resulting in complicated issues in data mining and ML. Imbalanced event datasets exhibit a prevalence of rare occurrences, wherein the quantity of instances associated with one class is significantly lower than the quantity of instances pertaining to the other. These datasets present challenges for learning algorithms since they may result in biased outcomes in downstream tasks such as classification, clustering, forecasting, and simulation. Algorithms necessitate tailoring to effectively address rare events, as these occurrences often give rise to challenges stemming from their uncommon nature.

Considering the significance of rare events across various domains, extensive research has been conducted in diverse knowledge areas, leading to a conflation of terms and issues within the literature. While terms like rare events, anomalies, novelties, and outliers may appear similar, it is crucial to note their distinct differences. Rare events and anomalies share characteristics such as an imbalanced class distribution and representation of all classes in the training set \cite{carreno2020analyzing}. However, rare events primarily relate to temporal data, contrasting with anomalies that typically involve static data distributions. Novelties, on the other hand, involve the identification of new or unknown data patterns. Static novelties are characterized by supervised classification with a single class for training, whereas dynamic novelties entail supervised classification with an indeterminate number of labels \cite{carreno2020analyzing, de2015evaluation, pimentel2014review}. Outliers, distinguished from the aforementioned categories, are observations significantly deviating from the majority in the dataset \cite{carreno2020analyzing, l2001outliers}. They often manifest in temporal data and are addressed through unsupervised classification methodologies. The primary focus of our investigation in this paper is exclusively directed towards rare events. Hence, we specifically selected studies that include the term  ‘rare events’  in the paper title and abstract.

In real-life, rare events can be observed ubiquitously in various domains, including medical diagnosis, fraud detection, and natural disaster prediction. In any field or area of study, rare events can be regarded as occurrences that possess valuable and meaningful information \cite{omar2022exploring}. Rare events can be weighted by `\textit{rarity}', a measure of being rare, uncommon, or scarce. Liu and Feng \cite{liu2022curse} formulate this as the "Curse of Rarity" (CoR). The fundamental idea behind CoR is that the events of interest are exceptionally rare, resulting in limited information in the available data. CoR leads to many underlying issues, including decision-making, modeling, verification, and validation. For instance, detecting rare diseases or medical conditions is challenging but crucial for effective diagnosis and treatment.
Similarly, detecting fraud in financial transactions can help prevent financial losses and ensure the security of transactions. In the case of natural disaster prediction, identifying rare events such as earthquakes or tsunamis can help in effective disaster management and response. In manufacturing, these events lead to unplanned downtime or shutdowns, which are particularly detrimental for industries regarding equipment life and power consumption. Thus, exploring rare events in advance allows industries to implement mitigation procedures to reduce defects so that equipment downtime can be lowered, optimizing energy consumption and ensuring optimization, quality, and safety standards in processes.

The significance of rare events resides in their capacity to yield a disproportionate influence, surpassing that of more typical events. For instance, a disease that affects a minor percentage of the population can enormously affect public health. Similarly, large-scale fraudulent activities can result in substantial financial losses for individuals and organizations. Hence, it is evident that rare occurrences require specialized attention and analysis. Therefore, developing effective methodologies and algorithms that can handle the uniqueness and mitigate the biases and limitations inherent in rare events is essential. The problem of imbalanced datasets and rare events is not new, and researchers have developed several techniques to address this issue. These techniques range from data-level approaches, such as oversampling and undersampling \cite{seiffert2007mining, zhao2018framework, ranjan2018dataset, jo2004class, wu2007local, ahmadzadeh2019rare}, to algorithm-level approaches, such as cost-sensitive learning \cite{zhao2018framework, li2017rare, kubat1998machine, nugraha2020clustering} and ensemble methods \cite{dai2016random}. In recent years, deep learning methods \cite{alestra2014rare, ashraf2023identification, olmucs2022comparison, bhanja2022black, ranjan2020understanding, martello2021improving, xu2022training, hsieh2019unsupervised} have also been applied to address the problem of rare events.

This survey paper aims to provide a comparative review of the existing literature on rare event prediction. We have developed a taxonomy that summarizes the research on rare event prediction into four categories: rare event data, data processing, algorithmic approaches, and evaluation approaches. The rare event data section identifies datasets containing rare events, while the data processing section emphasizes the vital role of data processing in handling rare event datasets, refining data quality to enhance predictive model performance. The algorithmic approaches present mathematical models applicable to diverse scenarios encompassing various use cases, and the evaluation section delves into the multifaceted evaluation criteria utilized in rare event prediction studies. Then, we explore the existing literature pertaining to these four categories. The extended review involves the reviewing of each of these four categories with levels of rarity that we introduce, established industries, types of datasets, data modalities, applications, and downstream tasks. Finally, we highlight some open research questions and future directions identified in this area.

\subsection{The Contributions of this paper}
The main contributions of this paper include:
\begin{enumerate}
\item 
Comparatively review the existing literature on rare event prediction in four approaches; rare event data, data processing, algorithmic and evaluation approaches.
\item 
Analyze the literature by examining multiple avenues: dataset types, modalities, and downstream tasks. 
\item 
Identify gaps, challenges, and special concerns in the current research landscape while discussing potential emerging trends in the realm.
\end{enumerate}

\subsection{Organization}
\begin{figure}[!ht]
  \centering
  \includegraphics[width=0.8\linewidth]{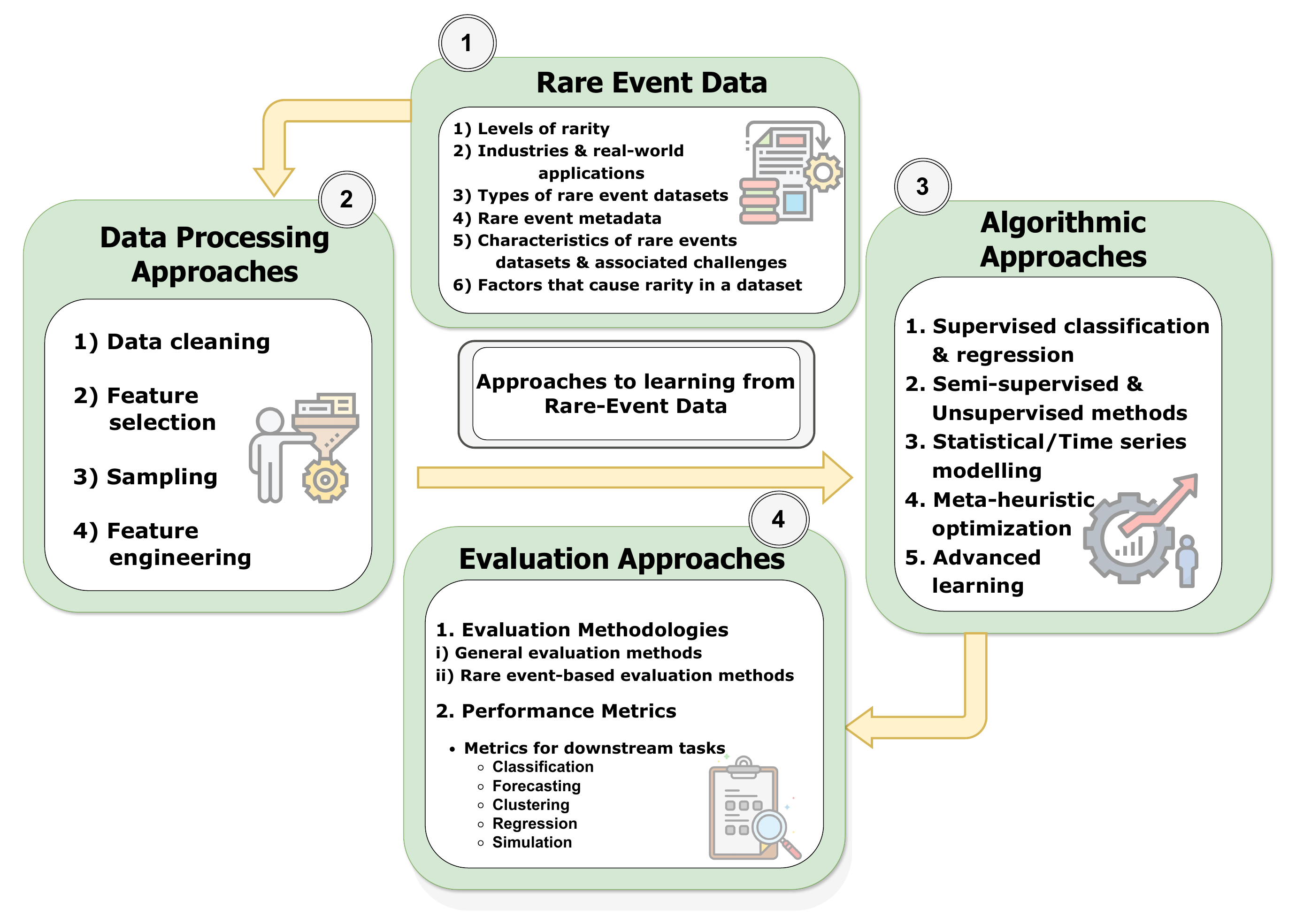}
  \caption{Approaches to learning from rare event data}
  \label{fig:rareeventapproaches}
\end{figure}

This study distinguishes a four-fold summarized categorization of approaches to learning from imbalanced event data for rare event prediction that have been implemented within related work. As shown in Figure \ref{fig:rareeventapproaches}, the main groupings include I) Rare event data, II) Data processing approaches, III) Algorithm level techniques, and IV) Evaluation approaches.

\clearpage

\section{Rare event data}
This section analyzes datasets with rare events. Considering a wide range of datasets from multiple industries and different modalities we first categorize them by the rarity percentage. We then explore real-world applications of the data and examine data acquisition methods. Next, we present an analysis considering types of datasets, metadata, and modality. Finally, we discuss the characteristics and challenges of handling such data and the factors contributing to their rarity.

\subsection{Datasets with rare events – Analysis of existing datasets with rare events}
\subsubsection{Levels of rarity}
\begin{figure}[h]
  \centering
  \includegraphics[width=0.7\textwidth]
  {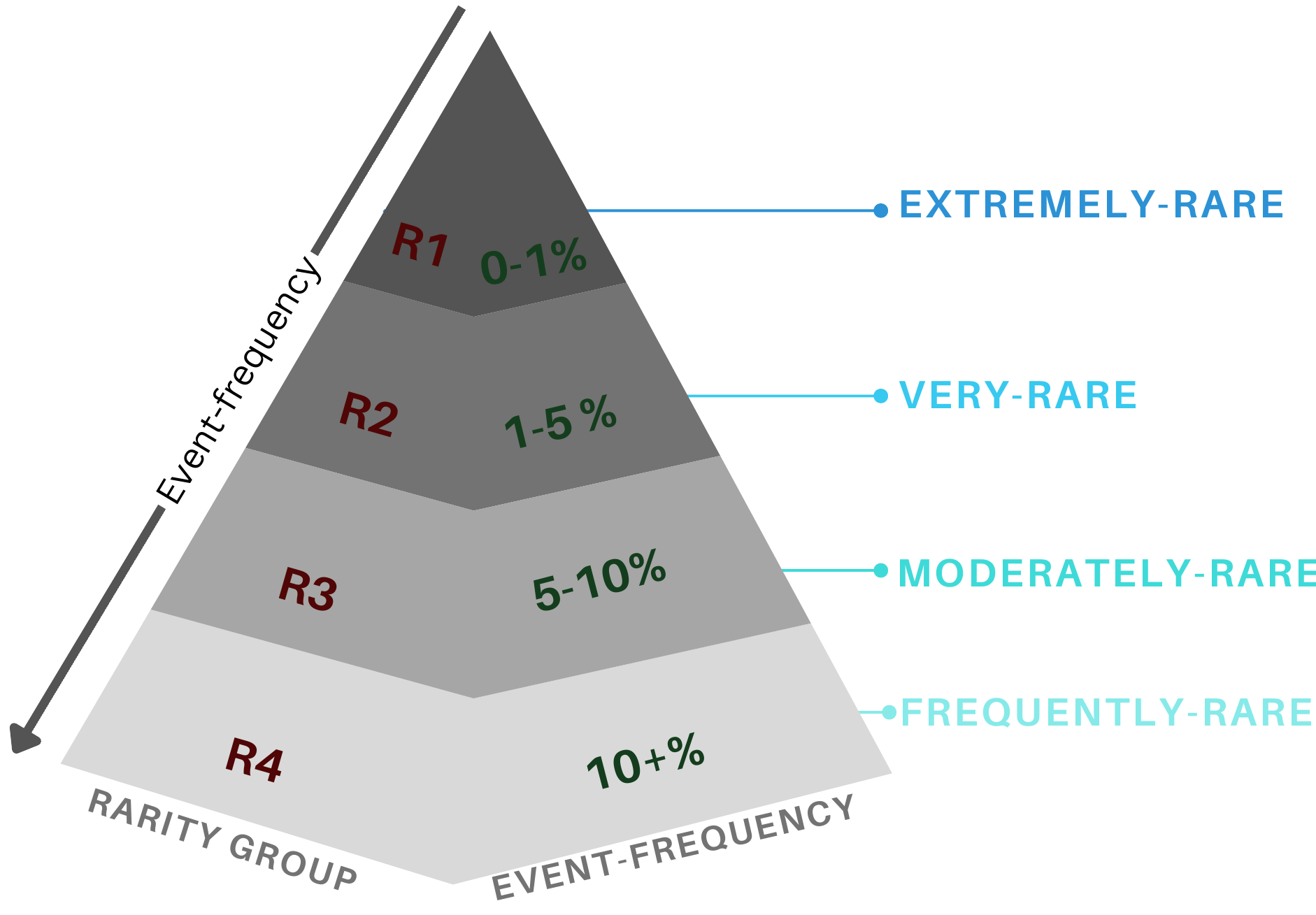}
  \caption{Levels of rarity}
  \label{fig:levelsofrarity}
\end{figure}

In any domain, the rarity of events is inversely correlated with the maturity of that industry. At the same time, rarity is correlated with event frequency or the probability of occurrence. For better understanding and analysis, we introduce the notion of \textit{``levels of rarity"}, which categorizes rarity into four levels as depicted in Figure \ref{fig:levelsofrarity}. These rarity levels will be used throughout all the review sections in this paper. We have established the boundaries for the levels based on several factors, such as the distribution of data, significant differences in rarity levels, and context. The R1 category comprises datasets with events that have a frequency of 0-1\%, which are considered extremely rare. On the other hand, events with a rarity of 1-5\% are classified as very rare and fall under the R2 category. The R3 category includes events with a rarity of 5-10\%, which are considered moderately rare. Finally, the rest of the events with a frequency greater than 10\% belong to the frequently-rare (R4) category. It is apparent that when moving up in levels, the percentage of rarity and the event frequency tend to decrease, necessitating more sophisticated approaches for identifying and analyzing. The datasets in the studies we reviewed utilized diverse data types, including numeric (N), textual (TX), image (I), and audio (A). A special property of these `rare event' datasets is their adherence to the temporal nature (T) of `events` and/or adaptation of time-dependent features. 
We have encountered a limited number of datasets (2) that deviate from the typical time series data used in rare event studies, which we have excluded from our analysis. Consequently, we delineate the scope of rare events as occurrences within a time series, where an entire or part of a time series may constitute a rare event evolving over an extended duration.

\subsubsection{Industries and real-world applications}
The different industries that have been identified in this review include eight main sectors: economy, healthcare, transportation, telecommunications, manufacturing, energy, earth science, and others. It should be noted that the primary industries include several application domains that we merged into a general main industry, as shown in Table \ref{tab:freq}. In Table \ref{tab:freq}, we have compiled a list of real-world applications of rare events categorized by industry, application domain, and rarity based on the literature. Notably, these applications have predominantly concentrated on use cases such as detection, diagnosis, prediction, and downstream tasks like classification (CF), clustering (CL), forecasting (FT), regression (RG), and simulation (SM). 

\begin{table}[!ht]
\centering
\caption{List of selected rare event applications}
\scriptsize
\label{tab:freq}
  \begin{tabular}{lll}
    \toprule
Industry  & Applications Domains         & Examples from literature \\
\midrule
Earth Sciences & \begin{tabular}[c]{@{}l@{}} Environment management, Disaster management,\\ Geology \end{tabular} & \begin{tabular}[c]{@{}l@{}}Detection of changes in buildings from aerial images taken before and after\\a tsunami disaster (\cite{fujita2017damage, hamaguchi2019rare} -- R1)\\$O_3$ state prediction, Prediction of hazardous seismic bumps in coal mines (\cite{li2017rare} -- R3)\\Detection of oil spills in satellite-borne radar images (\cite{kubat1998machine, marins2021fault} -- R2)\\Prediction of landslides, undesirable events in offshore oil wells (\cite{githubGitHubPetrobras3W} -- R4)\end{tabular} \\
\midrule

Manufacturing & Machinery fault diagnosis, Anomaly detection & \begin{tabular}[c]{@{}l@{}}Prediction of paper breaks in the paper manufacturing industry (\cite{ranjan2018dataset, Ranjan2019DataCD} -- R1)\\Prediction of which parts will fail at the end of the production line (\cite{hebert2016predicting} -- R1)\end{tabular} \\
\midrule

Telecommunication & \begin{tabular}[c]{@{}l@{}} Network monitoring, Telecommunication \\ failure diagnosis \end{tabular} & \begin{tabular}[c]{@{}l@{}}Prediction of telecommunications equipment failures (\cite{weiss1998learning, stateRoadwayData} -- R1)\\Prediction and classification of rare events in power grids (\cite{tamuIEEE39Bus, uciMachineLearningSpam} -- R1, R2, R3, R4)\end{tabular} \\ 
\midrule
Transportation    & \begin{tabular}[c]{@{}l@{}}Traffic monitoring, Autonomous driving, \\
Aircraft health management     \end{tabular}                                                                       & \begin{tabular}[c]{@{}l@{}}Degradation detection and trending,\\ and failure discrimination based on the classification of aircraft systems (\cite{alestra2014rare} -- R1)\\Prediction of wrong-way driving in highways (\cite{ashraf2023identification} --R1)\end{tabular}   \\  
\midrule
Economy & Market trend forecasting, Fraud detection & \begin{tabular}[c]{@{}l@{}}Prediction of black-swan events in the Indian stock market (\cite{bhanja2022black} --R1)\\Predicting credit card frauds (\cite{nugraha2020clustering} -- R1)\end{tabular} \\ 
\midrule
Healthcare        & \begin{tabular}[c]{@{}l@{}}Disease diagnosis, Emergent incidents identification, \\Detecting microcalcifications\end{tabular}               & \begin{tabular}[c]{@{}l@{}}Post-Traumatic Seizure Detection (\cite{ravindranath2020m2nn} -- R1)\\Detection of LASA cases and rare events’ classification in imbalanced \\ healthcare data (\cite{zhao2018framework} --R4) \\Detection of microcalcifications in mammography (\cite{chen2004using} -- R2)\end{tabular}                                                                          \\   \midrule
Energy  &   Energy forecasting  & Prediction of when to expect a high KWh cost (\cite{berberidis2007detection} --R3)    \\  \midrule
Others            & \begin{tabular}[c]{@{}l@{}}Scene understanding, Computer vision, Remote sensing, \\ Classification, Criminological investigation\end{tabular} & \begin{tabular}[c]{@{}l@{}}Detection of scene changes from street view panorama images (\cite{hamaguchi2019rare} -- R1)\\Detection of recidivists from criminological events (\cite{huang2004learning} -- R4)\\Classification of types of glass in criminological events (\cite{wang2012applying, kubat1997addressing} -- R4)\\Detection of a change of digit from a pair of samples in MNIST (\cite{hamaguchi2019rare} -- R1, R4)
\end{tabular}      \\ 
\bottomrule
\end{tabular}
\end{table}

\subsubsection{Types of rare events datasets}
The study identified three types of rare event datasets: naturally rare event datasets, derived datasets, and simulated/synthetic datasets. 
\\

\textbf{\textit{i) Naturally rare event datasets (RE)}}:
These refer to datasets that inherently exhibit a low occurrence rate of specific events or phenomena. Table \ref{tab:naturalraredatasets} summarizes naturally rare event datasets used in rare event literature. It was noticed that approximately 41\% of the naturally rare datasets used fall into the extremely-rare category. Naturally rare event datasets can be sourced from various industries and domains, such as manufacturing, healthcare, earth sciences, and economy. For instance, data on environmental disasters such as oil spills \cite{kubat1998machine} and tsunamis \cite{fujita2017damage} can be gathered using satellite, ariel, or drone imagery via remote sensing technologies, while machine faults and anomalies can be tracked through data collection from sensors attached to different parts of the machines in manufacturing plants \cite{ranjan2018dataset, case_school_of_engineering_2021}. Similarly, sources for predicting rare economic events may include financial market data, economic indicators, government reports, transaction data, consumer behavior data, social media data, and other relevant sources specific to the economic domain \cite{bhanja2022black}. Data collected from such sources can be obtained from public databases. Repositories like University of California Irvine (UCI) \cite{uciMachineLearning}, Knowledge Extraction based on Evolutionary Learning (KEEL) \cite{ugrKEELSoftware}, Outlier Detection DataSets (ODDS) \cite{Rayana_2016}, Kaggle \cite{kaggleFindOpen}, research-specific data storages \cite{githubGitHubPetrobras3W, case_school_of_engineering_2021, tamuIEEE39Bus, harvardSWANSF,  APSFailureDataSet, PHM_Society}, industry-specific databases \cite{stateRoadwayData, skybraryAircraftCondition, ecmwfAccessForecasts, noaaStormPrediction, noaaStormData, noaaStormEvents, nasaIntelligentSystems, fdotCerberusClient, fdotCerberusClient2022}, Application Programming Interfaces (APIs) \cite{googleGoogleMaps, noaaServicesversion, nasaNASAOpen}, social media platforms \cite{bseindiaformerlyBombay, nse_india}, and news outlets \cite{bseindiaformerlyBombay, nse_india} would be such databases. Data extraction from them can involve manually accessing public datasets, querying databases, web scraping, or using web services.

\begin{table}[!ht]
\footnotesize
\caption{Naturally rare datasets.$^*$}
\label{tab:naturalraredatasets}
\begin{tabular}{llll}
\toprule
Sector & 
\begin{tabular}[c]{@{}l@{}}
Event \% \& \\ rarity group 
\end{tabular} 
& Datasets with modality & Papers  \\
\midrule
\\

Earth Sciences & 0-1(R1)                  & \begin{tabular}[c]{@{}l@{}}meteorological (heatwaves) dataset(N,T), \\ 3W dataset(N,T) \cite{githubGitHubPetrobras3W}\end{tabular}                                                                                                                                                                                                                                    & \cite{berberidis2007detection,vargas2019realistic}                                                                                                             \\
                                & 1-5(R2)                  & \begin{tabular}[c]{@{}l@{}}Oil dataset(I, T) \cite{kubat1998machine}, \\ 3W dataset(N,T) \cite{githubGitHubPetrobras3W},\\ Tornado dataset(N,T) \cite{lakshmanan2005neural}\end{tabular}                                                                                                                                            & 
                                \begin{tabular}[c]{@{}l@{}}
                                \cite{chen2004using, kubat1997addressing, kubat1998machine}, \\ \cite{vargas2019realistic, tang2008svms, maalouf2011robust} \end{tabular}                                             \\
                                & 5-10(R3)                 & \begin{tabular}[c]{@{}l@{}}seismic-bumps Data Set\\ (N, T) \cite{uciMachineLearningcoal}, 3W dataset(N, T) \cite{githubGitHubPetrobras3W}\end{tabular}                                                                                                                                                                                              & \cite{li2017rare, vargas2019realistic}                                                                                                                         \\
                                & 10+(R4)                  & \begin{tabular}[c]{@{}l@{}}LIDAR data(I, T)\cite{van2009combined}, 3W dataset(N,T) \cite{githubGitHubPetrobras3W}\end{tabular}                                                                                                                                & \cite{van2006prediction, vargas2019realistic}                                                                                                                  \\ \\
Manufacturing & 0-1(R1)                  & \begin{tabular}[c]{@{}l@{}}pulp -and-paper dataset(N,T) \cite{ranjan2018dataset},\\ 
Bosch Production Line \\ Performance dataset(N,T) \cite{kaggleBoschProduction} \end{tabular}                                                                                                                                                                        & 
\begin{tabular}[c]{@{}l@{}}
\cite{Ranjan2019DataCD, hebert2016predicting, ranjan2018dataset, ravindranath2020m2nn}, \\ \cite{neuman2021extreme, lee2021early, martello2021improving, xu2022training} \end{tabular} \\
                                & 10+(R4)                  & \begin{tabular}[c]{@{}l@{}}Case Western Reserve University \\Rolling Bearing dataset(N,T) \cite{case_school_of_engineering_2021},\\ \\ IMS bearing dataset)(N,T) \cite{qiu2006wavelet},\\ XJTU-SY datasets(N,T) \cite{wang2018hybrid},\\ PRONOSTIA bearing dataset(N,T) \cite{nectoux2012pronostia}\end{tabular} & \cite{rehab2021bearings, liu2021machinery, xu2021two, wang2018hybrid}                                                                                          \\ \\
Telecommunication               & 0-1(R1)                  & \begin{tabular}[c]{@{}l@{}}probe binary-UCI(N, T),\\ r2l binary-uci(N,T ),\\ KDD Cup 99 Data (N,T) \cite{Rayana_2016, uci1999Data, uciMachineLearningKDD99,kaggle1999Data}\\ Alarm data(n,T),\\ VoIP traffic data(N, T) \cite{stateRoadwayData}\end{tabular}                                                                                                      & \cite{wu2007local, weiss1998learning, meng2006rare, maalouf2018logistic}                                                                                       \\ \\
Transportation & 1-5(R2)                  & \begin{tabular}[c]{@{}l@{}}Air Pressure System(APS) Failure \\at Scania Trucks Data Set(N, T)\\ \cite{uciMachineLearningTrucksAPS, APSFailureDataSet}\end{tabular}                                                                                                                                                                       & \cite{gondek2016prediction, fathy2020learning, rafsunjani2019empirical}                                                                                        \\
                                & 5-10(R3)                 & \begin{tabular}[c]{@{}l@{}}MnDot traffic data(N, T) \cite{stateRoadwayData}\\ Traffic Prediction Dataset(N, T)\\ WWD Data (N, T) 
\cite{fdotCerberusClient, fdotCerberusClient2022}\end{tabular}                                                                                                                                                    & \cite{ashraf2023identification, meng2006rare, ravindranath2020m2nn}                                                                                            \\ \\
Economy                         & \begin{tabular}[c]{@{}l@{}}
0-1(R1)   \end{tabular}               & \begin{tabular}[l]{@{}l@{}}S and P BSE SENSEX(N, T)\cite{nse_india, kaggleYearSensex}\\  Nifty 50(N, T) \cite{bseindiaformerlyBombay, kaggleNIFTY50Stock}\\ Kaggle Credit Card Fraud Detection(N, T)\end{tabular}                                                      & \cite{bhanja2022black, nugraha2020clustering}   \\ \\

Healthcare & 0-1(R1)                  & \begin{tabular}[c]{@{}l@{}}Thoracic surgery dataset(N, T) \cite{li2017adaptive}, \\ Bioassay AID 746(N, T),687(N, T),456(N, T),373(N, T), \\ Suicide data(N,T) \cite{githubGitHubMHResearchNetworkDiagnosisCodes} \end{tabular} 

& \cite{li2017adaptive,coley2023empirical}                                                                                                                     \\ \\
                                & 1-5(R2)                  & \begin{tabular}[c]{@{}l@{}}stroke dataset(N, T) \cite{kaggleStrokePrediction},\\ Bioassay AID 362(N, T) \cite{li2017adaptive}\end{tabular}                                                                                                                                                                                                                 & \cite{omar2022exploring, li2017adaptive}                                                                                                                       \\
                                & 5-10(R3)                 & Bioassay AID 1608(N, T) \cite{li2017adaptive}                                                                                                                                                                                                                                                                                                                            & \cite{li2017adaptive}                                                                                                                                          \\
                                & 10+(R4)                  & Wong’s dataset from Canadian(TX ,T) \cite{wong2016statistical}                                                                                                                                                                                                                                                                                                           & \cite{zhao2018framework, wong2016statistical}                                                                                                                  \\ \\
Energy                          & 5-10(R3)                 & 
\begin{tabular}[c]{@{}l@{}}Daily electric energy production \\ measurements dataset(Spain, 2003)(N,T)    
\end{tabular}
& \cite{berberidis2007detection}                                                                                                                                 \\

Others          & 0-1(R1)                  & PCD dataset(I, T) \cite{jst2015change},                                                                                                                                                                                                                                                                                                                               & \cite{hamaguchi2019rare}                                                                                                                                       \\
                                & 1-5(R2)                  & K1b-WebACE(N,T) \cite{han1998webace}                                                                                                                                                                                                                                                                                                                                  & \cite{wu2007local}                                                                                                                                             \\
                                & 5-10(R3)                 & La12-TREC(N,T)                                                                                                                                                                                                                                                                                                                                                                         & \cite{wu2007local}                                                                                                                                             \\
                                & 10+(R4)                  & \begin{tabular}[l]{@{}l@{}}Recidivism dataset (N, T) ,\\ Audio-Anomaly-Dataset(A, T) \cite{kaggleAudioAnomalyDataset}\end{tabular}                                                                                                                                                                                                                                          & \cite{huang2004learning, abbasi2022large}                                                                                                                      \\

\bottomrule              
\end{tabular} \\
\footnotesize{$^*$N-Numeric, TX-Textual, I-Image, A-Audio, T-Time series}
\end{table}

\textbf{\textit{ii) Derived datasets (DE)}}:
A derived dataset is a dataset that is not initially rare, results from transformation of an existing dataset, and includes a new insight, `derived rarity'. For this, the original dataset should ideally possess sufficient information and features related to multiple events of interest and should initially be not rare. The availability of such data enables performing operations, calculations, or algorithms on the raw dataset to transform it into a derived dataset that captures the `derived rarity' aspect \cite{li2017rare, alestra2014rare, dangut2022application}. One example of derived datasets is the ABCD (AIST Building Change Detection) dataset \cite{fujita2017damage, githubGitHubGistaircABCDdataset, abciABCDDataset}, which was initially not designed to include rare events but has been manipulated in certain studies to create rare event scenarios by altering event percentages \cite{ahmadzadeh2019rare, hamaguchi2019rare, cheon2009bayesian}. MNIST dataset \cite{googleMNISTVariations}, originally an image dataset, has been repurposed in studies to detect letter changes, creating an imbalanced dataset by reducing the frequency of certain letters \cite{ahmadzadeh2019rare}. Table \ref{tab:deriveddatasets} summarizes derived datasets used in rare event research in the studied literature. The analysis findings indicate that a substantial portion (34\%) of the derived datasets we considered in this review belong to the extremely rare category. 

\begin{table}[!ht]
\footnotesize
\caption{Derived datasets.$^*$}

\label{tab:deriveddatasets}
\begin{tabular}{llll}
\toprule
Sector                             & 
\begin{tabular}[c]{@{}l@{}}
Event \% \& \\ rarity group  

\end{tabular}
& Papers                                                                           & Source Datasets with modality \\
\midrule

Earth Sciences  & 0-1(R1)                   & \cite{hamaguchi2019rare, cheon2009bayesian, ahmadzadeh2019rare} & \begin{tabular}[c]{@{}l@{}}ABCD (AIST Building Change Detection) dataset(I, T) \cite{fujita2017damage, githubGitHubGistaircABCDdataset, abciABCDDataset}, \\  Weather data(N,T),Air pollutant data(N,T),\\ Space Weather ANalytics for Solar Flares (SWAN-SF) \\ benchmark dataset(N, T) \cite{harvardSWANSF}\end{tabular} \\ \\
                                   & 1-5(R2)                   & \cite{ahmadzadeh2019rare}                                       & 
                                   \begin{tabular}[c]{@{}l@{}}
                                   ABCD (AIST Building Change Detection) dataset(I, T) \cite{fujita2017damage, githubGitHubGistaircABCDdataset, abciABCDDataset}, \\ Space Weather ANalytics for Solar Flares (SWAN-SF) \\  benchmark dataset(N,T) \cite{harvardSWANSF}                       \end{tabular}    \\ \\
                                   & 5-10(R3)                  & \cite{kubat1997addressing, cheon2009bayesian}                   & \begin{tabular}[c]{@{}l@{}}Oil dataset(I,T) \cite{kubat1998machine}, \\ Weather data(N, T), Air pollutant data(N,T)\end{tabular} \\  \\
                                   & 10+(R4)                   & \cite{cheon2009bayesian, murthy2007distributed}                 & \begin{tabular}[c]{@{}l@{}}Weather data(N, T),Air pollutant data(N,T),\\ ABCD (AIST Building Change Detection) dataset(I,T) \cite{fujita2017damage, githubGitHubGistaircABCDdataset, abciABCDDataset},\end{tabular} \\  \\ 
Telecommunication & 0-1(R1)                   & \cite{he2021weighting, wu2007local, joshi2001mining}            & \begin{tabular}[c]{@{}l@{}}IEEE 39-bus power system data(N,T) \cite{tamuIEEE39Bus},\\ KDD Cup-99 (N,T) \cite{uci1999Data, uciMachineLearningKDD99,kaggle1999Data}\end{tabular}   \\ \\
                                   & 1-5(R2)                   & \cite{he2021weighting}                                          & IEEE 39-bus power system data(N, T) \cite{tamuIEEE39Bus}\\  \\
                                   & 5-10(R3)                  & \cite{he2021weighting}                                          & \begin{tabular}[c]{@{}l@{}}IEEE 39-bus power system data(N, T) \cite{tamuIEEE39Bus},\\ Spam data(N, T) \cite{uciMachineLearningSpam}\end{tabular}  \\ \\
                                   & 10+(R4)                   & \cite{he2021weighting}                                          & \begin{tabular}[c]{@{}l@{}}IEEE 39-bus power system data(N, T) \cite{tamuIEEE39Bus},\\ Spam data(N, T) \cite{uciMachineLearningSpam}\end{tabular} \\  \\ 
Transportation                     & 0-1(R1)                   & \cite{dangut2022application, alestra2014rare}                   & \begin{tabular}[c]{@{}l@{}} AIRBUS data(N, T),\\  ACMS dataset(N, T) \cite{skybraryAircraftCondition}\end{tabular}          \\  \\

Healthcare        & 0-1(R1)                   & \cite{ravindranath2020m2nn}                                     & EEG Seizure Dataset(N, T)               \\ \\
                                   & 1-5(R2)                   & \cite{xiu2021variational, ravindranath2020m2nn}                 & \begin{tabular}[c]{@{}l@{}}COVID-19(N, T),\\ InP -Duke University Health System (DUHS)(N,T) \cite{o2020development}, \\ SEER(N, T) \cite{ries2007seer},\\ EEG Seizure Dataset (N, T)\end{tabular}     \\ \\
                                   & 5-10(R3) & \cite{xiu2021variational}                      & COVID-19(N, T)      \\ \\

Energy           & 
\begin{tabular}[c]{@{}l@{}}
0-1(R1), \\
1-5(R2),\\ 
5-10(R3), \\
10+(R4)
\end{tabular}
& \cite{bai2022rare}                             & MAGIC Gamma Telescope(N,T) \cite{ucimagicgamma}                                   \\
Others             & 0-1(R1)                   & \cite{hamaguchi2019rare}                                        & \begin{tabular}[c]{@{}l@{}}Augmented MNIST(I,T) \cite{googleMNISTVariations}, \\ WDC dataset(I,T) \cite{hamaguchisupplementary}\end{tabular}                                                                                                                                                                             \\  \\
                                   & 10+(R4)                   & \cite{seiffert2007mining, li2017rare, hamaguchi2019rare}        & \begin{tabular}[c]{@{}l@{}}Augmented MNIST(I,T) \cite{googleMNISTVariations},\\ Adult dataset(N,T) \cite{uci_adult},
                                   \\  AudioSet dataset(A,T) \cite{audioset}
                                   
                                   \end{tabular}                     \\  \\ 
 \bottomrule   

 \end{tabular}
 \\
 \footnotesize{$^*$N-Numeric, TX-Textual, I-Image, A-Audio, T-Time series}
\end{table}

\begin{table}[!ht]
\centering
\footnotesize
\caption{Simulated \& Synthetic datasets.$^*$}
\label{tab:sim_syn_datasets}
\begin{tabular}{lllll}
\toprule
Sector                          & \begin{tabular}[c]{@{}l@{}} Event \% \& \\ rarity group \end{tabular} & Papers     & Data type   & Technique   \\                                                                          
\midrule
Earth Sciences & 0-1 (R1)                  & \cite{marins2021fault, vargas2019realistic, pickering2022discovering}  & N, T                                                                            & \begin{tabular}[c]{@{}l@{}} OLGA Dynamic Multiphase \\ Flow Simulator \cite{slbOLGADynamic}, MATLAB  \end{tabular}\\
                                & 1-5 (R2)                  & \cite{marins2021fault, vargas2019realistic}                                                                                                                                                                                   & N, T                                                                            & \begin{tabular}[c]{@{}l@{}} OLGA Dynamic Multiphase \\ Flow Simulator \cite{slbOLGADynamic}, MATLAB  \end{tabular}\\
                                & 5-10 (R3)                 & \begin{tabular}[c]{@{}l@{}}\cite{cheon2009bayesian, kubat1997addressing}\\ \cite{marins2021fault, vargas2019realistic}\end{tabular} & N, T                                                                            & 
                                \begin{tabular}[c]{@{}l@{}} MATLAB, OLGA Dynamic Multiphase \\ Flow Simulator \cite{slbOLGADynamic},   \end{tabular}\\

                                & 10+ (R4)                  & \cite{marins2021fault, vargas2019realistic}                                                                                                                                                                                   & N, T                                                                            & 
                                \begin{tabular}[c]{@{}l@{}}OLGA Dynamic Multiphase \\ Flow Simulator \cite{slbOLGADynamic}        \end{tabular} \\
                                &
                                \begin{tabular}[c]{@{}l@{}}
                                Rarity not \\ reported
                                \end{tabular} 
                                & \cite{chalongvorachai2021data, chalongvorachai20213dvae, kaewkiriya20223dvae} 
                                & N, T &
                                \begin{tabular}[c]{@{}l@{}}
                                 Signal Fragment Assembler (SFA), \\ Variational Autoencoder (VAE), \\ Data Picker (DP), \\ 
Quality Classifier (QC) \end{tabular} \\

                                \\

Others          & 0-1 (R1)                  & \begin{tabular}[c]{@{}l@{}}\cite{olmucs2022comparison, ling2021support, dhulipala2022active, bai2022rare,joshi2001mining} \end{tabular}                                           & \begin{tabular}[c]{@{}l@{}}N, T\\ N, T\end{tabular} & \begin{tabular}[c]{@{}l@{}}Monte Carlo,  MATLAB\\ \\ Importance sampling\end{tabular}  \\
                                &  \begin{tabular}[c]{@{}l@{}}
                                1-5 (R1), \\ 5-10 (R3) 
                                \end{tabular} & \cite{dhulipala2022active, bai2022rare}                                                                                                                                                                                       & \begin{tabular}[c]{@{}l@{}}N, T\\ N, T\end{tabular} & \begin{tabular}[c]{@{}l@{}}Monte Carlo,  MATLAB\\ 
                                
                                Importance sampling\end{tabular}  \\
                                & 10 + (R4)                & \cite{li2017rare}                                                                                                                                                                                                             & N, T                                                                            & MOA \cite{bifet2010moa} \\             \bottomrule  
\end{tabular}\\
\footnotesize{$^*$N-Numeric, T-Time series}
\end{table}

\textbf{\textit{iii) Simulated / Synthetic (SIY) datasets}}:
A simulated (also termed synthetic) dataset is an artificially generated dataset that mimics the characteristics and patterns of real-world rare events. They are typically created based on known models, distributions, or algorithms to replicate the statistical properties and relationships observed in the original data. These data can be generated in cases where collecting direct data on rare events is challenging and impractical. They are beneficial for predicting rare events that have not yet occurred or for testing the accuracy of predictive models in a controlled environment \cite{olmucs2022comparison, marins2021fault, githubGitHubPetrobras3W, kubat1997addressing, cheon2009bayesian}. In the literature, SIY datasets have been generated in controlled environments using tools such as MATLAB, OLGA Dynamic Multiphase Flow Simulator \cite{slbOLGADynamic}, Signal Fragment Assembler (SFA), Variational Autoencoder (VAE), Data Picker (DP), Quality Classifier (QC) \cite{chalongvorachai2021data, chalongvorachai20213dvae, kaewkiriya20223dvae}, and frameworks like Massive Online Analysis (MOA) \cite{li2017rare, bifet2010moa}, as presented in Table \ref{tab:sim_syn_datasets}. While most studies focus on naturally rare and derived datasets, considerably less research is based on simulated / synthetic datasets.

\subsubsection{Rare event metadata}
Metadata refers to the descriptive information that provides additional context about a dataset. It includes information about the structure, format, quality, source, and other data characteristics. In rare event data acquisition, metadata acquisition involves extracting relevant information from the data that contributes to a better understanding of the rare events. While metadata may be applicable to both rare and majority classes, its importance for rare events stems from the need to capture additional attributes specific to these events. This can include attributes such as timestamps, geographical location, variables related to the event’s occurrence, data sources, data collection methods, and any other contextual information that aids in analyzing and interpreting the rare events. The following methods were revealed as ways of acquiring metadata.

\begin{enumerate}[i]
    \item Data collection: Metadata can be collected during data acquisition. This includes capturing information such as the source of the data, timestamps, geographical coordinates, sensor settings, or any other relevant contextual details that help describe the data \cite{bhanja2022black, dangut2022application,ben2022roc, kaupp2021context}.

    \item Expert opinions: Engaging domain experts or subject matter specialists can provide valuable metadata. Experts can contribute their knowledge about rare events, their causes-effects, associated variables of interest, equations, hypotheses, or factors that influence their occurrence \cite{cheon2009bayesian, dangut2022application, olmucs2022comparison, vargas2019realistic, marins2021fault, githubGitHubPetrobras3W}. This knowledge can assist in identifying appropriate metadata to enhance the analysis and prediction of rare events.

    \item Data annotation: Adding annotations or labels to the data can serve as metadata. This requires manually categorizing data instances as rare events or non-rare events, assigning event severity or damage levels, utilizing standard metrics and indexes, and providing additional descriptive labels that capture specific attributes or characteristics of the rare events \cite{olmucs2022comparison, yang2022prediction, bhanja2022black, adil2022deep, vargas2019realistic, berberidis2007detection}.

    \item External sources: Documents, technical reports, publications, and websites offer insights, statistical data, or contextual details that contribute to the understanding and analysis of rare events. For example, clinical reports, insurance claims data, state mortality records from government websites, outpatient visit details from electronic health records, and responses to patient-reported measures like the Patient Health Questionnaire (PHQ-9) have provided valuable information in predicting rare medical incidents \cite{coley2023empirical, simon2018predicting}. 

\end{enumerate}

\subsubsection{Characteristics of rare event datasets and associated challenges}
Rare event datasets exhibit distinct characteristics that lead to various related issues, some outlined below.
\begin{enumerate}[i]

    \item Skewed class distribution and lack of data: A skewed class distribution is a distribution of classes that is not symmetrical or evenly distributed. Class imbalance is a specific case of skewed class distribution featuring a substantial disparity in the number of instances between classes \cite{wang2018systematic, monard2002learning}. In rare events datasets, the minority class has a significantly smaller number of samples than the majority class. This skewness in class distribution makes it difficult for ML algorithms to learn patterns and classify the minority class accurately \cite{kubat1998machine}. Lack of data can take two forms. Absolute rarity occurs where the number of samples associated with the minority class is small in the absolute sense, whereas relative rarity happens where the minority class samples are less relative to the other classes \cite{weiss2004mining}. These rarity forms pose consequential challenges for classifiers in identifying patterns and regularities within these rare occurrences to learn a robust model \cite{kubat1998machine, olmucs2022comparison}.

    \item Temporal property: An inherent characteristic of rare event datasets is the temporal aspect, which refers to the occurrence or sequencing of events over time. Temporal property is essential in analyzing and understanding rare event data, as it provides insights into the timing, duration, order, and interdependencies of rare events and normal events \cite{carreno2020analyzing}. However, due to the imbalanced class distribution and data sparsity in a temporal context, capturing the time-dependent patterns and correlations in data can be a major challenge. Thus, handling these issues in a temporal context adds complexity to the analysis, necessitating specialized techniques for accurate prediction.

    \item Class overlap: In some cases, there may be overlapping patterns or similarities between the rare event(minority class) and the non-rare events(majority class) \cite{napierala2016types, stefanowski2013overlapping}. This can lead to misclassifying rare events as more common, resulting in false negatives.

    \item Uncertainty: Uncertainty, in the context of rare event datasets, refers to the lack of precise knowledge and confidence in the observed data. This arises from the limited sample size, data sparsity, high-class imbalance, and lack of information, and it becomes a challenging issue in generalizing the findings of any downstream task \cite{ bhanja2022black, dhulipala2022active}.

    \item High dimensionality: Rare event datasets can include many features or dimensions, including numerical, categorical, or textual variables \cite{gondek2016prediction, rafsunjani2019empirical}. Including multiple features complicates the modeling process and requires careful feature selection or dimensionality reduction techniques.

    \item Event complexity: Rare events involve intricate relationships between multiple variables, which generate complex patterns and interactions in real-world systems \cite{kaupp2021context}. This often makes modeling rare events difficult, necessitating sophisticated modeling approaches to capture the underlying complexity. 

\end{enumerate}

\subsubsection{Factors that cause rarity in a dataset}
Several factors can cause rarity in a dataset, including:

\begin{enumerate}[i]

\item Natural occurrences: Some events simply occur less frequently in nature than others. For example, diseases that affect a small percentage of the population will naturally result in highly imbalanced datasets. Natural hazards like landslides, tsunamis, and seismic bumps or climate catastrophes like heavy flooding that may occur once in a century are also rare occurrences \cite{li2017rare,van2006prediction, yang2022prediction, zhao2021event}. Black-swan events, unpredictable events with severe global economic consequences, are extremely rare natural events in the economic sector. Events like the 9/11 attack and the Chinese economic downturn are examples of these \cite{bhanja2022black}.

\item Class definition: The definition of a class can also impact its rarity. If a class is defined narrowly or specifically, it may contain fewer instances than if defined broadly \cite{napierala2016types}. For instance, when estimating the likelihood of stroke based on varied input parameters like demographic data, additional illnesses, and smoking status, the definitions of "having a stroke" and "not having a stroke" may be narrowly defined. Due to these specific criteria used in classification, stroke events within the dataset may be deemed rare.

\item Sampling bias: If data are collected from a biased sample, they may not precisely reflect the true distribution of the population. In particular, if a rare event dataset only contains events occurring in certain geographic regions or in certain types of people, this can cause rarity. For instance, while HIV (Human Immunodeficiency Virus) is not classified as rare due to its global prevalence \cite{WHO2022}, there are specific regions and subpopulations where HIV is relatively rare, indicating variations in its incidence and prevalence.

    \item Cost and measurement errors: Data collection and labeling can be expensive, and in some cases, collecting large amounts of data for a rare event may not be feasible. Evidently, the generation of simulated data and synthetic data may be costly. It may be challenging to accurately observe and measure rare events in natural and controlled environments, resulting in fewer labeled instances. Similarly, rare events that involve complex interactions among their dimensions may not always be observable and measurable accurately. For example, rare events like Wrong-Way Driving (WWD) are influenced by multiple variables like roadway geometry and configurations,  traffic volume, lighting conditions, weather conditions, and driver's age and medical conditions. However, the impact of these complex interactions is not always readily observable and measurable, giving rise to various complications \cite{ashraf2023identification}.

    \item Subjective decisions: Curators can selectively include instances of a rare event based on specific criteria by selecting features and classes that impact the rarity of that class within the dataset. The decision to include only such instances can affect the rarity of the class within the dataset. This is apparent when creating new datasets by generating derived, simulated, or synthetic data based on the original rare event datasets. For example, in predicting rare events like extreme heat waves, curators have used subjective decisions by selecting temperature and humidity thresholds and time periods based on their understanding of heat wave severity \cite{berberidis2007detection}. These decisions can be made to categorize the data into classes and identify patterns related to moderate, severe, or extreme heat waves.
\end{enumerate}

\begin{figure}[!ht]
  \centering
  \includegraphics[width=1\linewidth]{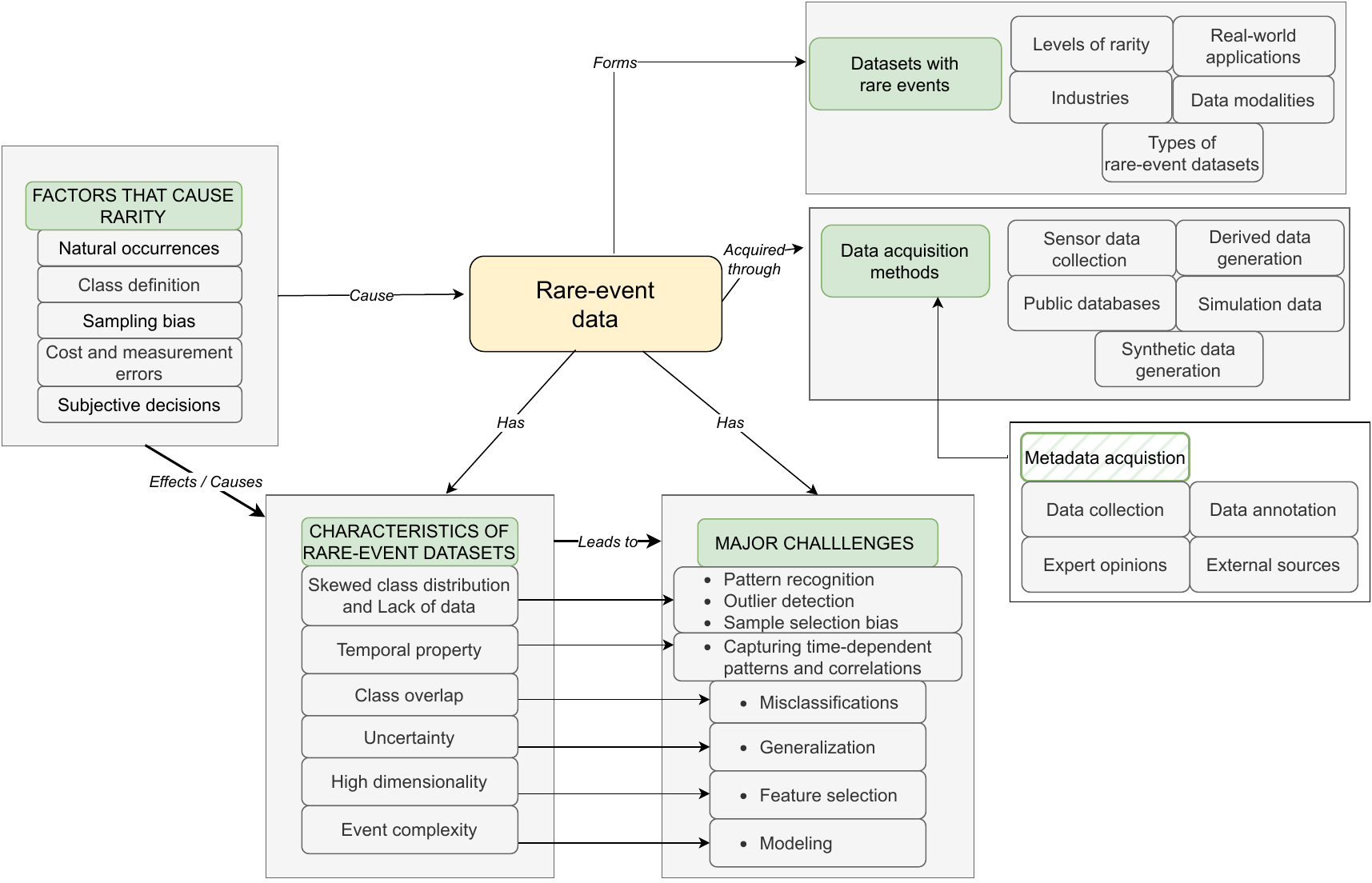}
  \caption{Relationship between rare event data, acquisition methods, rarity factors, characteristics, and challenges of rare event datasets}
 
  \label{fig:overalldatarelationship}
\end{figure}

\noindent The relationship between rare event data, acquisition methods, rarity factors, characteristics, and challenges of rare event datasets is multifaceted and interconnected, as summarized in Figure \ref{fig:overalldatarelationship}. The factors contributing to a dataset's rarity inherently result in rare event data and their characteristics. Rare event data constitute rare event datasets, which can be acquired through various methods. The characteristics of these data, in turn, give rise to various challenges in analyzing and predicting rare events, as discussed in the previous section.

In conclusion, this section examined four significant subsections in depth: rare event datasets, data acquisition methods, factors of rarity, characteristics, and challenges of dealing with rare event data. We devised a rarity hierarchy that provided a systematic method for summing data pertinent to rare events for the purposes of analysis. While most datasets and studies fall under the extremely-rare category in the hierarchy of rarity, many research projects are based on naturally rare and derived datasets. Textual and audio data-based research on rare events has received less attention than time series, image, and numerical data-based datasets. It is worth revealing that rare event-related problems and research are not restricted to a specific domain, industry, or sector, as we investigated research efforts across multiple sectors. Finally, the relationship between rare event data, acquisition methods, rarity factors, characteristics, and challenges of rare event datasets are drawn to summarize the overall scope of review in this section.

\section{Data Processing Approaches}
This section focuses on the importance of data processing methods in enhancing data quality for improved predictive model performance when dealing with rare event datasets. It explores various data processing approaches used in rare event prediction research, emphasizing their specific objectives. The subsequent discussion provides a detailed examination of each objective, including an analysis of how these approaches intersect with data modality, rarity groups, and downstream tasks.

\subsection{Objectives of Data Processing Approaches}
\begin{figure}[!ht]
  \centering
  \includegraphics[width=1\linewidth]{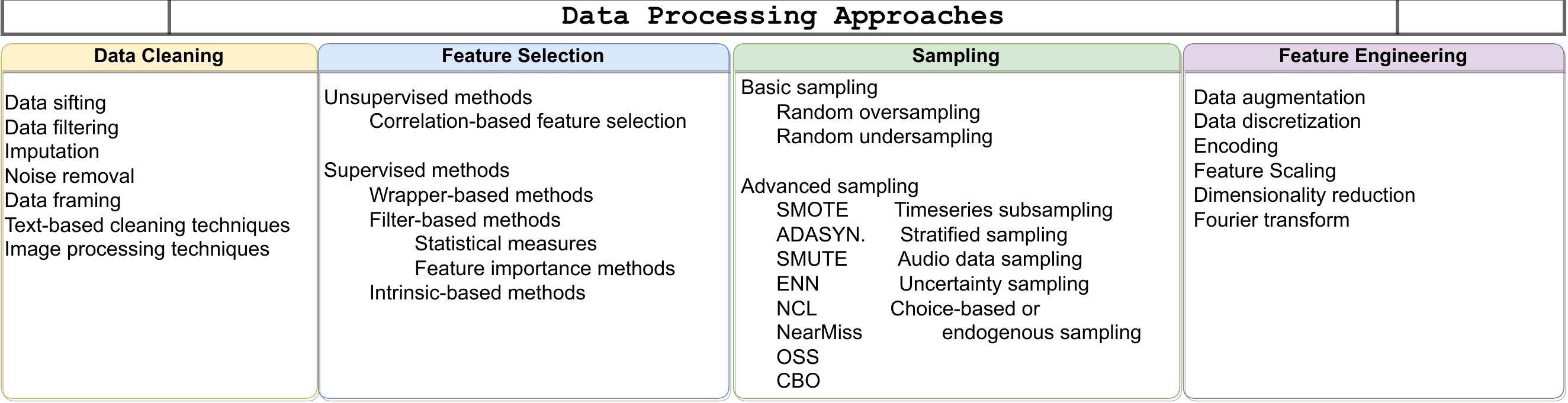}
  \caption{Data processing approaches in rare event research}
  \label{fig:dataprocessing}
\end{figure}

In the literature, we identified that data processing approaches aim to achieve four main objectives. Firstly, it's responsible for data cleaning, which improves the quality, consistency, and reliability of data analysis results. Secondly, it caters to feature selection by selecting the optimal variables by limiting the input variables to the model and utilizing only relevant features. Thirdly, it aids in sampling by modifying the data samples to balance the distribution and/or eliminating undesirable samples at the data level. Finally, data processing methods are applied in feature engineering to transform raw series data into a stable format suitable for modeling. In Figure \ref{fig:dataprocessing}, we classify the data processing approaches in rare event research that adhere to the aforementioned four objectives into four main categories. The approaches to data processing in rare event studies are summarized in Table \ref{tab:dataprocessing}, along with their primary categories, rarity groups, data modalities, and subsequent tasks.

\subsection{Data Cleaning (DC)}
Data cleaning focuses on the important task of preparing and refining the data. This involves various techniques and processes to improve the dataset's quality, consistency, and reliability. This subsection examines various data cleaning approaches and discusses the application of them in various data modalities, rarity levels, and downstream tasks (See Figure \ref{fig:datacleaning} and discussion in the sub section \ref{sec:dc1}.

\subsubsection{Approaches to data  cleaning}
Various approaches address data cleaning tasks, often depending on the specific data modalities. Some common data cleaning methods used in rare event-based data mining include data sifting, filtering, imputation, noise removal, audio-based preprocessing, text-based preprocessing, and image processing techniques.

\subsubsection{Approaches to data  cleaning}
Various approaches address data cleaning tasks, often depending on the specific data modalities. Some common data cleaning methods used in rare event-based data mining include data sifting, filtering, imputation, noise removal, audio-based preprocessing, text-based preprocessing, and image processing techniques.

\textbf{\textit{i) Data sifting}}:
Refers to refining large volumes of data to identify the most relevant and important information. In rare event literature, sifting has been used to systematically sort through the data to identify and extract specific subsets of data. There are two types of data sifting: heuristic data sifting and statistical data sifting. Heuristic data sifting relies on expert knowledge and intuition, while statistical data sifting relies on quantitative measures and algorithms to identify relevant features \cite{meng2006rare, cheon2009bayesian, lan2010study, gujrati2007meta}. In \cite{cheon2009bayesian}, important knowledge about Ozone ($O_3$) concentration (i.e., time periods where $O_3$ is high) has been used as a heuristic, and several statistical measures were undertaken to identify relationships between different $O_3$ states and related components. 

\textbf{\textit{ii) Data filtering}}:
Involves segregating and removing unwanted, irrelevant data or information from a large dataset. This includes techniques like removing duplicate data \cite{hsieh2019unsupervised}, removing records containing very small number of records per group \cite{ xu2022training, xu2022new}, removing rows containing irrelevant types \cite{meng2006rare}, and removing rows that meet specific conditions \cite{xiu2021variational}. For instance, in \cite{meng2006rare}, only the connected calls were used, discarding the unconnected calls from VoIP traffic data in unsupervised rare event detection in spatiotemporal environments. \cite{xiu2021variational} used a cut-off time to define an event of interest and excluded subjects censored before the cut-off time.

\textbf{\textit{iii) Imputation}}:
Imputation or value approximation techniques estimate and represent missing and incorrect values in a dataset with reasonable approximations based on the available data. These techniques are vital for addressing the sparse, incomplete, and imbalanced nature of the rare event data, requiring careful consideration to avoid skewed results. Simple methods like mean and median imputation, though commonly used, can introduce bias in rare event scenarios due to the skewed nature of the data \cite{APSFailureDataSet, xiu2021variational, rafsunjani2019empirical, fathy2020learning, gondek2016prediction,uciMachineLearningTrucksAPS}. Advanced techniques such as Interpolation techniques \cite{radi2015estimation, ahmadzadeh2019rare}, Iterative Imputation \cite{stekhoven2012missforest, adil2022deep,omar2022exploring}, Multiple Imputation by Chained Equation(MICE)\cite{rafsunjani2019empirical}, Soft Impute \cite{yao2018accelerated}, Expectation maximization (EM) \cite{do2008expectation,omar2022exploring}, and Singular Value Decomposition (SVD) \cite{wei2018missing, troyanskaya2001missing} are better suited for rare-event datasets, as they account for complex patterns and correlations \cite{adil2022deep, yao2018accelerated, wei2018missing}. These methods help maintain the integrity of rare events during the imputation process, making them significant for accurate predictions.

\textbf{\textit{iv) Noise removal}}: Involves eliminating random and irrelevant data variations that don't contribute meaningful information to patterns and relationships. The Brooks-Iyengar algorithm is a fault-tolerant method for sensor fusion, handling faulty sensor readings effectively  \cite{brooks1996robust}. Iyer et al. \cite{iyer2015statistical} utilized Brooks-Iyengar along with ensemble data cleaning trees to handle random noise and missing data. Sampling methods like Tomek-links (TL) and Edited Nearest Neighbors (ENN) are used for noise removal. TL removes noise and boundary points in the majority class during rare events  \cite{kotsiantis2006handling, li2017rare, ashraf2023identification, kubat1998machine}, while ENN eliminates noisy data samples, resulting in smoother decision boundaries \cite{ashraf2023identification}.

\textbf{\textit{v) Text-based cleaning techniques}}: Standard text preprocessing techniques, such as converting text to lowercase, removing stop-words, and employing stemming and lemmatization, have been used in rare event prediction studies involving textual data. This standardization improves the reliability of predictions by addressing textual inconsistencies, reducing noise and data variability in rare events involving textual data  \cite{wong2016statistical, zhao2018framework}. For instance, in healthcare, R packages such as 'tm', 'snowball', and 'Rstem' have been used to standardize text data by normalizing word variations through stemming and lemmatization, thereby enhancing the prediction of rare medical incidents \cite{wong2016statistical}.

\textbf{\textit{vi) Image processing techniques}}:
In rare event prediction using images, image processing techniques have been employed to clean and normalize the images. These techniques aim to detect suspicious regions and extract relevant features that differentiate rare regions from similar ones. Kubat et al. \cite{kubat1998machine} used various image processing methods to correct for radar beam incidence angle, identify dark regions, and extract specific features like the size of oil spills and average brightness. The output of the image processing is a fixed-length feature vector for each suspicious region, facilitating further analysis and prediction.

\subsubsection{Analyzing data cleaning approaches with data modalities, rarity groups, and downstream tasks}
\label{sec:dc1}

\begin{figure}[!ht]
  \centering
  \includegraphics[width=0.8\linewidth]{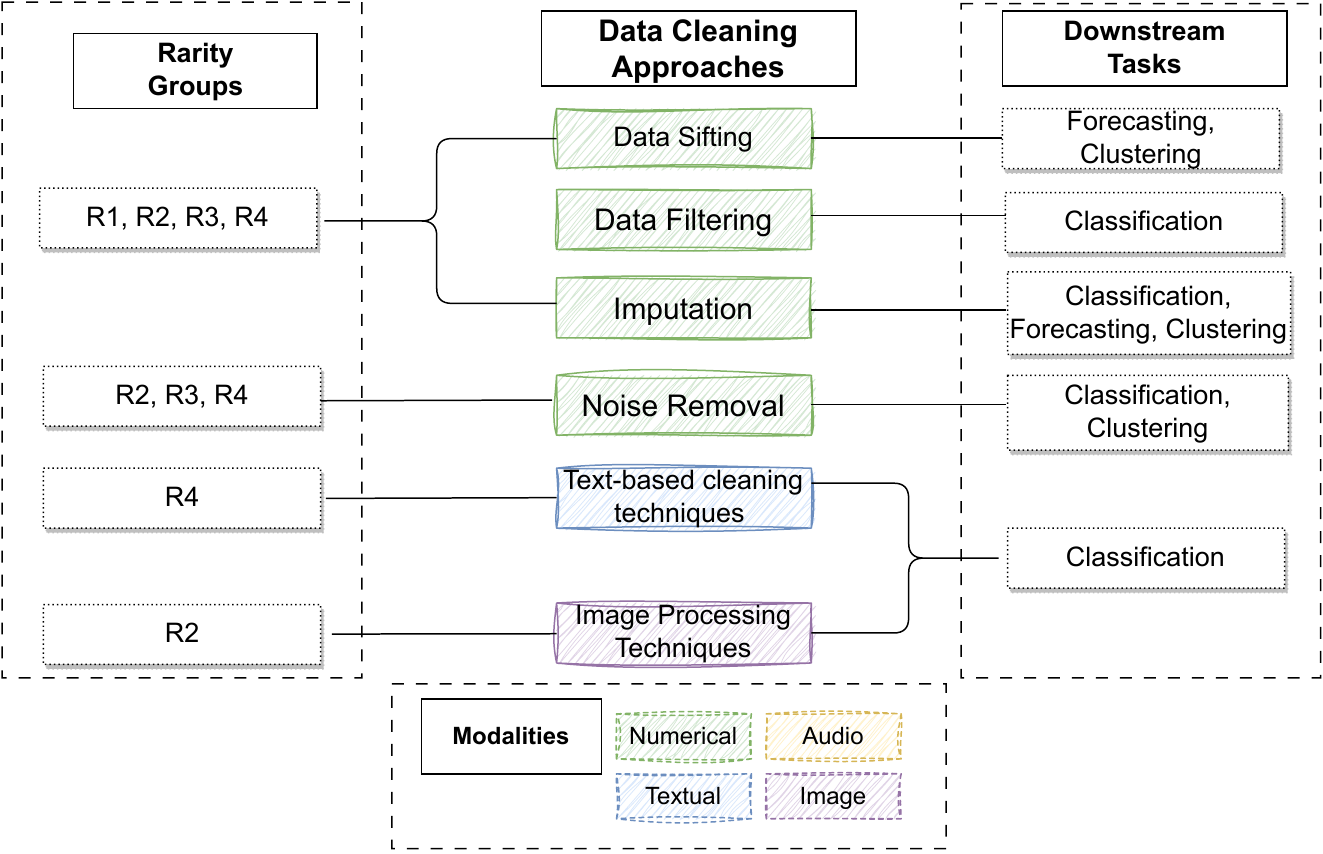}
  \caption{Association between data cleaning approaches, data modalities, rarity groups and downstream tasks \\
* Coloring of data cleaning approaches corresponds to the data modalities}
  \label{fig:datacleaning}
\end{figure}

To explore the data cleaning approaches, we referred to 116 rare event prediction-related papers. Then, we analyzed these by data cleaning approach,  modality, and rarity group. We observed the interplay between these as shown in Figure \ref{fig:datacleaning}. In terms of numerical data cleaning, techniques such as data sifting, data filtering, imputation, and noise removal have been commonly employed, with the rarity level being independent of these techniques. Notably, noise removal has not been applied to the extremely-rare group. Moreover, specific modalities are addressed, such as image processing techniques for image-based rare event prediction and text-related techniques encompassing textual summary generation, text conversions, stemming, and lemmatization for text-based predictions. These audio-based and text-based approaches were used for frequently-rare datasets. Furthermore, image processing techniques are employed to predict images within the very-rare group. While many data processing methods have supported classification tasks, some of these methods have been used in clustering and forecasting research.

\subsection{Feature Selection (FS)}
Feature selection aims to identify a subset of input features from a dataset to extract the most pertinent information and improve the model's predictive capacities by reducing complexity. This subsection examines various approaches for feature selection in rare event studies and discusses the application of these techniques to various data modalities, rarity levels, and downstream tasks (See Figure \ref{fig:featureselection} and discussion in Section \ref{sec:fc1}).

\subsubsection{Approaches to feature selection}
Feature selection in rare event prediction is crucial due to the imbalance and sparsity of the data, requiring careful adaptation to identify the most relevant features while preserving critical information related to rare events. The techniques can be categorized into unsupervised and supervised methods.
\\ \\
\textbf{\textit{I) Unsupervised methods}}:
Unsupervised feature selection methods are applied without considering the response or target variable, focusing on the relationships and patterns within the independent variables. For example, Correlation-based feature selection, frequently used in rare-event studies, evaluates the relationships between features using correlation coefficients like Pearson correlation, Analysis of Variance (ANOVA), and chi-squared tests. However, given the rarity of the events, thresholds for eliminating features must be set carefully to preserve variables that may appear redundant in general datasets but are critical in rare-event contexts \cite{li2017rare, dangut2022application, olmucs2022comparison, mogos2023distribution}.
\\ 

\textbf{\textit{II) Supervised methods}}:
These methods use the response/target variable in the feature selection process and eliminate irrelevant variables in making a prediction. The supervised methods can be categorized into wrapper-based, filter-based, and intrinsic-based. Wrapper-based methods search for well-performing subsets of input features. They wrap or embed an ML algorithm within their core to perform feature selection. In rare-event scenarios, wrapper-based methods like Recursive Feature Elimination (RFE) can be adapted to preserve features critical for identifying rare events, even if these features do not show strong significance in a broader dataset. For instance, RFE has been combined with Hidden Markov Models (HMMs) to predict rare events, as it is sensitive to the subtle temporal patterns that may indicate these uncommon occurrences \cite{rehab2021bearings}.
Standard filter-based methods extract or select features based on their statistical relationship with the target variable. Given their rarity, traditional statistical measures can be misleading or insufficient to identify rare events. For instance, sliding window techniques and wavelet transforms can be adjusted to focus on detecting rare spikes or patterns that could be easily overshadowed by more frequent but less relevant data \cite{xu2021two}. The adaptation of methods like wavelet analysis \cite{xu2021two}, Discrete Wavelet Transform (DWT) \cite{rehab2021bearings} and Gumbel copula function \cite{yang2022prediction}, Minimum Redundancy Maximum Relevance (mRMR) \cite{mogos2023distribution}, Term Frequency-Inverse Document Frequency (TF-IDF) \cite{wong2016statistical, zhao2018framework} as seen in rare-event studies involves fine-tuning these techniques to capture the unique, low-frequency signals that often characterize rare events. Furthermore, statistical methods ranging from simple calculations like mean, median, variance, skewness, kurtosis, and standard deviation to more advanced measures like spectral energy and frequency entropy derived from time and frequency domain analyses have been used in rare event prediction in the financial domain \cite{bhanja2022black,neuman2021extreme}. In rare-event prediction, feature importance methods require special attention to ensure that features relevant to rare occurrences are not overlooked. Methods like Gini importance \cite{hebert2016predicting, gondek2016prediction, omar2022exploring} or XGBoost’s accuracy and cover measurements \cite{hebert2016predicting} have been adapted to emphasize features that, while they might contribute minimally to overall model accuracy, are critical in predicting rare events. For example, the application of Mel-frequency Cepstral Coefficients (MFCC) in rare-event contexts goes beyond traditional audio signal processing, aiming to capture subtle acoustic variations that could signify the occurrence of an event, as demonstrated in studies focusing on audio-based rare-event prediction \cite{abbasi2022large, shi2020few}.
Intrinsic methods, such as attention mechanisms, are particularly valuable in rare-event prediction due to their ability to dynamically focus on the most relevant features across different scales and dimensions.  For instance, multi-variate and multi-scale attention methods have been specifically adapted to enhance the prediction of rare events by focusing on spatial-temporal features that are often subtle and dispersed \cite{ravindranath2020m2nn}. Additionally, decision trees, which are intrinsically capable of feature selection, have been adapted to handle the unique distributional characteristics of rare-event datasets, ensuring that the rare but critical branches of the tree are not pruned away during model training \cite{li2017rare, kubat1998machine}.

\subsubsection{Analyzing feature selection approaches with data modalities, rarity groups, and downstream tasks}
\label{sec:fc1}

\begin{figure}[!ht]
  \centering
  \includegraphics[width=0.8\linewidth]{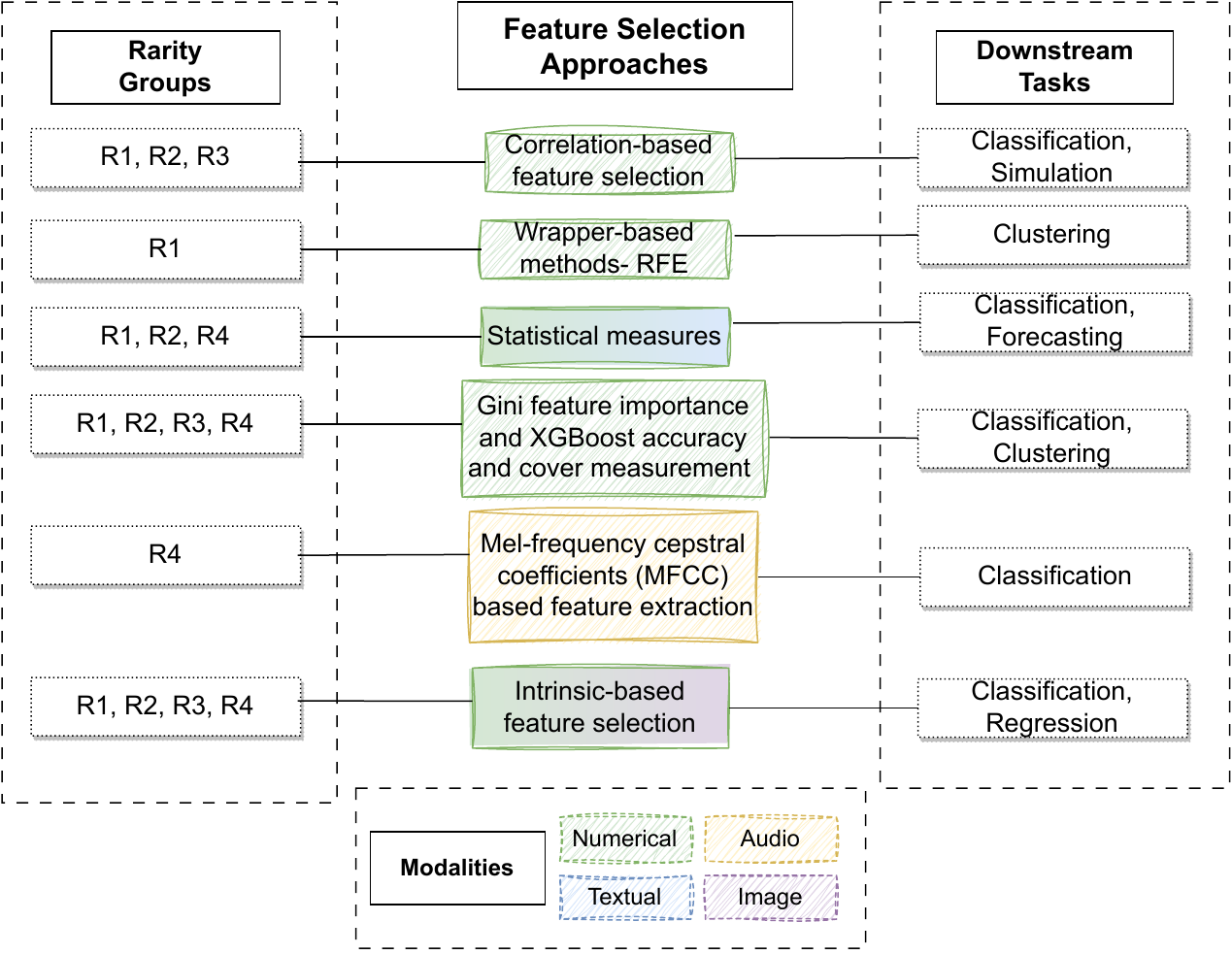}
  \caption{Association between feature selection approaches, data modalities, rarity groups and downstream tasks \\
  * Coloring of feature selection approaches corresponds to the data modalities
}
  \label{fig:featureselection}
\end{figure}

Figure \ref{fig:featureselection} illustrates the intricate association between feature selection methods, modality variations, and rarity groups within the context of rare event prediction, as analyzed across the reviewed papers. Regarding numerical-based feature selection, feature importance, and intrinsic-based methods are independent of the rarity groups. Correlation-based feature selection, wrapper-based, filter-based, and intrinsic methods have been used with data belonging to extremely-rare and very-rare groups.  TF-IDF and MFCC-based feature extraction has been used with frequently-rare data. Decision tree-based intrinsic methods were utilized with very-rare image datasets.  Likewise, in data cleaning, most data processing methods support classification tasks; some have been used in clustering, regression, simulation, and forecasting research.

\subsection{Sampling (SL)}
Sampling techniques in ML are essential methods for selecting a subset of data from larger datasets and are often used to tackle challenges like class imbalance and large data sizes while enhancing computational efficiency. The application of rare event data showcases their effectiveness in significantly improving performance in mining rare events, highlighting their importance in research contributions. This subsection examines various approaches for sampling and discusses their application to various data modalities, rarity levels, and downstream tasks (See Figure \ref{fig:sampling} and discussion in Section \ref{sec:sl1}).

\subsubsection{Approaches to sampling}
Sampling techniques used in rare event prediction can be categorized into basic and advanced methods based on the complexity and sophistication of the sampling approaches utilized.

\noindent \textbf{\textit{I) Basic sampling techniques}}: 
These techniques seek to address the issue of class imbalance by eliminating instances from the majority class or increasing the minority class by duplicating minority class samples. Random minority oversampling (ROS) and random majority undersampling (RUS) are the most frequent basic sampling strategies. In ROS, instances of the minority class are replicated randomly in the dataset, while in RUS, occurrences of the majority class are randomly eliminated from the dataset. In rare event-based research, many studies have used ROS and RUS methods to achieve a more balanced class distribution \cite{seiffert2007mining, zhao2018framework, ranjan2018dataset, jo2004class, wu2007local, ahmadzadeh2019rare}. Some researchers have combined sampling techniques with clustering models \cite{nugraha2020clustering, wu2007local} and ensemble learning methods \cite{chen2004using} to improve predictive performance. Some others have applied randomly over/under-sampling in conjunction with advanced architectures, such as the Siamese CNN \cite{hamaguchi2019rare}, to enhance rare event detection. Additionally, statistical sampling methods, like Hoeffding bounds, have been employed in rare event learning using associative rules and higher-order statistics \cite{iyer2015statistical}. Even though basic sampling has been extensively used, it has several limitations. The drawback of the oversampling method is that it leads to overfitting since the model learns from the same duplicated samples repeatedly. Undersampling eliminates lots of data that could have been utilized to train the model and improve its accuracy.

\noindent \textbf{\textit{II) Advanced sampling techniques}}: 
Going beyond basic random adjustments, advanced sampling techniques utilize intelligent mechanisms that consider the distribution of data points and the nuance of learning specific examples, resulting in greater effectiveness for handling complex, imbalanced datasets. 

\textbf{\textit{1) Synthetic minority oversampling technique (SMOTE)}}:
SMOTE is a popular advanced sampling technique used in ML to address class imbalances. It generates new synthetic minority cases by extrapolating from existing minority instances, and it considers the difference between a sample and its closest neighbor to create synthetic examples. SMOTE has been adopted in various rare event-related use cases implementing different modeling techniques \cite{zhao2018framework, li2017adaptive, ali2014dynamic, ashraf2023identification, fathy2020learning}. It has been observed that logistic regression combined with SMOTE produces good results in detecting Look-Alike-Sound-Alike (LASA) cases in textual data \cite{zhao2018framework}, but it can be computationally expensive. Adaptive swarm balancing algorithms \cite{li2017adaptive} and dynamic churn prediction frameworks  \cite{ali2014dynamic} have also utilized SMOTE for rare event prediction in imbalanced datasets.
Additionally, Borderline-SMOTE is a variation that selectively applies SMOTE to minority instances on the border of the minority decision region, yielding effective results in mining tasks \cite{seiffert2007mining}. While SMOTE has been adopted in various rare event use cases, it has a significant drawback due to the arbitrary generation of synthetic data. As a result, the class boundaries between the majority class and the minority class in the synthetic data may appear significantly different from those in the original dataset, potentially deviating from the actual distribution of the minority class. 

\textbf{\textit{2) Adaptive synthetic sampling (ADASYN)}}:
ADASYN \cite{he2008adasyn} is a method that solves the issue with SMOTE by following a weighted distribution for minority classes according to their level of difficulty in learning. It generates synthetic observations of the harder-to-learn minority samples compared to the easier-to-learn minority samples and adaptively shifts the decision boundary towards the harder-to-learn samples. Asraf et al. \cite{ashraf2023identification} have used ADASYN with the XGBoost model in rare event modeling in highrisk WWD roadway segment identification. 

\textbf{\textit{3) Similarity majority undersampling (SMUTE)}}: SMUTE \cite{li2017rare} is an undersampling technique that distinguishes between the majority and minority class samples by considering the cosine similarity between each and its neighboring minority class samples. SMUTE works by calculating similarity scores between each majority class sample and a given number of minority class samples, then selecting a subset of majority class samples with the highest percentage of high similarity scores based on a specified undersampling rate.

\textbf{\textit{4) Edited nearest neighbor (ENN)}}:
ENN \cite{tomek1976experiment} is an undersampling method that reduces noise and refines decision boundaries in imbalanced datasets. ENN eliminates samples whose class labels differ from most of their nearest neighbors, resulting in a dataset with smoother decision boundaries and reduced noise. It has been applied with SMOTE oversampling to create a balanced, noise-free training dataset for improved rare event prediction \cite{li2017rare, ashraf2023identification}.

\textbf{\textit{5) Neighborhood cleaning rule (NCL)}}: NCL is an undersampling technique that removes redundant, noisy, or ambiguous samples. NCL employs the ENN technique to remove the data samples \cite{bekkar2013imbalanced}. However, compared with ENN, NCL has the additional benefit of removing redundant instances based on feature space similarity \cite{li2017rare}.

\textbf{\textit{6) NearMiss (NM)}}:
NM \cite{mani2003knn} is a technique that uses distance measures of majority class samples to minority class samples in selecting samples. When two points from different classes are located very close to one another in a distribution, this algorithm eliminates the data point from the larger class to balance the distribution. NearMiss-2(NM2), a variation of NM, has been used in rare event prediction data processing \cite{li2017rare, ashraf2023identification}.

\textbf{\textit{7) One-sided selection (OSS)}}:
OSS is an undersampling technique that balances imbalanced datasets by removing redundant and noisy majority-class instances. It combines the use of Tomek Links to identify ambiguous points on the class boundary with the Condensed Nearest Neighbor (CNN) rule \cite{hart1968condensed} to eliminate distant redundant examples from the majority class, resulting in a minimally consistent subset without compromising model performance. OSS has the advantage over other methods of intelligently locating and removing redundant and noisy majority class instances by utilizing Tomek Links and the CNN rule. This makes a more refined and representative subset of the majority class. OSS has been employed in various rare event studies to address the high imbalance in learning tasks \cite{kubat1997addressing, kubat1998machine, chen2004using, seiffert2007mining}. 

\textbf{\textit{8) Cluster-based oversampling (CBO)}}:
CBO employs resampling by clustering each class's training data separately and then performing random oversampling, cluster by cluster \cite{jo2004class, seiffert2007mining}. CBO employs separate clustering of each class's training data followed by random oversampling within each cluster, enhancing the representation of rare classes in a balanced manner. The advantage of this method is that it considers both between-class and within-class imbalances and then oversamples the data to correct both imbalances simultaneously. 

\textbf{\textit{9) Time series subsampling}}:
This involves selecting a subset of time points from the original to create a new, shorter, and more balanced time series \cite{fukuchi1999subsampling, combes2022time}. The difference between subsampling and time series subsampling is that subsampling is a general term that refers to selecting a random subset of data points, while time series subsampling is a specific type of subsampling used for time series data involving the selection of a subset of time points from a time series. In industrial-based research on machinery fault diagnosis \cite{liu2021machinery}, subsampling of time series has been performed to consider the balanced time series length of normal and different bearing fault types. Further, time series subsampling reduces the computational cost of analyzing long time series and aids in training ML models.

\textbf{\textit{10) Stratified sampling}}:
It is a general sampling technique that divides the dataset into subgroups (strata) based on the target variable's classes and then randomly samples from each stratum to ensure representation of all classes. To sample normal and different bearing fault failure types, \cite{liu2021machinery} uses stratified techniques to represent these subgroups in the final dataset properly.

\textbf{\textit{11) Audio data sampling}}:
Data framing has been used as a data sampling technique for audio data \cite{abbasi2022large} that involves converting the audio data into a machine-readable format to fix the audio file sampling (frame) rate. In this method, the audio data is divided into frames, each representing a segment of the audio signal sampled at a specific rate. The total number of frames can be calculated by multiplying the sampling rate by the audio file's duration.

\textbf{\textit{12) Uncertainty sampling}}:
It is a data sampling technique widely used in active learning, particularly relevant to rare event prediction research. It entails selecting instances from a dataset based on the uncertainty or low confidence of their predicted labels by a ML model. The goal is to prioritize sampling data points for which the model lacks certainty in its predictions. This approach is valuable in scenarios where labeling data is resource-intensive, such as in rare event prediction \cite{pickering2022discovering}, as it allows researchers to actively select the most informative instances for annotation, thereby improving the model's performance with limited labeled data.

\textbf{\textit{13) Choice-based or endogenous sampling}}:
Endogenous sampling is a rare event prediction method that selects samples based on the dependent variable (y) rather than the independent variable (X). It aims to obtain a representative sample that accurately reflects the distribution of rare events in the dataset, addressing the imbalance issue and improving predictive performance. Choice-based or endogenous stratified sampling has been used in various applications, such as Light Detection and Ranging (LIDAR) maps, where non-landslide cells are sampled one to five times more than landslide cells to achieve better representation \cite{van2006prediction}. Variations of regression models have been experimented with in endogenous sampling approaches \cite{maalouf2018logistic}.

\subsubsection{Analyzing data sampling approaches with data modalities, rarity groups, and downstream tasks}
\label{sec:sl1}

\begin{figure}[!ht]
  \centering
  \includegraphics[width=0.8\linewidth]{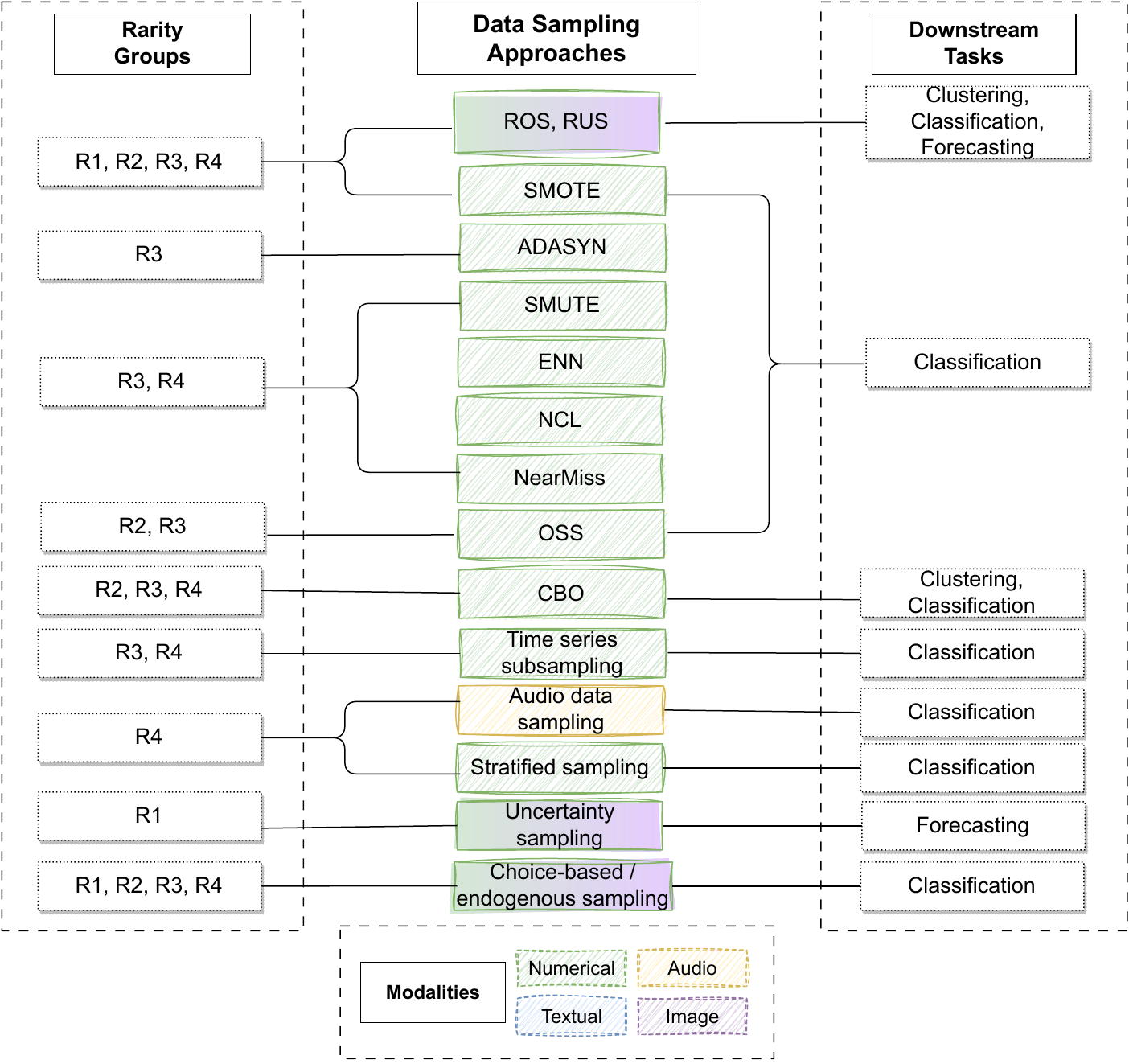}
  \caption{Association between sampling approaches, data modalities, rarity groups and downstream tasks \\
  * Coloring of sampling approaches corresponds to the data modalities
}
\end{figure}

Investigating sampling approaches, we analyzed 116 rare event prediction-related papers based on sampling approach, modality, and rarity group and observed their interrelationships as depicted in Figure \ref{fig:sampling}. It's seen that basic data sampling techniques are independent of rarity groups. They are used for numeric, image, and audio data and have been utilized in studies focusing on downstream tasks such as clustering, classification, and forecasting. The advanced sampling technique, SMOTE, is independent of rarity groups and is used in classification-based studies and numerical data. Most of the advanced sampling techniques, like SMUTE, ENN, NCL, NM, OSS, CBO, and time series subsampling, have been widely used in rare event prediction research with numeric datasets of varying rarity levels and are mostly used in classification-based studies. Uncertainty sampling and choice-based or endogenous sampling methods have been employed for regression and forecasting tasks across different rarity groups.

\subsection{Feature Engineering (FE)}
Feature engineering involves converting raw data into a relevant feature set for modeling. It aims to extract meaningful information from the dataset and present it in a format suitable for the learning algorithm. This subsection examines various approaches to feature engineering in rare-event contexts. Then, it discusses their application to various data modalities, rarity levels, and downstream tasks (See Figure \ref{fig:featureengineerning} and discussion in Section \ref{sec:fe1}).

\subsubsection{Approaches to Feature engineering}
Investigating rare events, we will review commonly utilized techniques in feature engineering as outlined below.

\textbf{\textit{I) Data augmentation}}:
In rare event prediction, data augmentation has been important in addressing the scarcity of event samples and the challenges of imbalanced datasets. Unlike general applications, where augmentation primarily boosts generalization, in rare events, it must be tailored to enhance model sensitivity to infrequent patterns. For instance, in the pulp-and-paper industry, a tailored augmentation method combined with Fast Fourier Transform (FFT) has been used to enrich time series data with synthetic rare event examples, improving fault detection \cite{Ranjan2019DataCD, ranjan2018dataset}. Wasserstein GANs (WGANs) have been adapted for rare event prediction by generating high-quality synthetic data to address data scarcity and improve training stability, while Conditional GANs (CGANs) are used in creating targeted samples based on specific attributes or conditions \cite{fathy2020learning}. Apart from those, standard image-based augmentation techniques, like cropping, have been applied to create additional training samples while preserving the critical spatial features necessary for detecting rare mineral deposits, predicting mineral prospectivity, and identifying scene changes \cite{yang2022applications, hamaguchi2019rare, parsa2021deep}. 
These targeted approaches ensure that models remain sensitive to rare events, improving their prediction accuracy.

\textbf{\textit{II) Data discretization}}: 
Data discretization can manage the unique challenges of skewed class distributions and high-dimensional data. Converting continuous data into discrete categories can help highlight the rare event patterns that might otherwise be overshadowed in continuous data \cite{omar2022exploring}, simplifying the model's ability to detect and learn from rare occurrences. Standard techniques like optimal binning have allowed supervised discretization that preserves essential characteristics of rare event data, ensuring that the model captures the subtle distinctions necessary for accurate prediction \cite{neuman2021extreme, berberidis2007detection}.

\textbf{\textit{III) Feature scaling}}: 
In rare event research, standardization has been used to align feature distributions with a standard normal distribution, which helps in mitigating the effect of class imbalance by ensuring that features do not disproportionately influence the model based on their scale \cite{liu2021machinery, lee2021early, xu2022training, abbasi2022large}. Normalization rescales values to a uniform range, preventing features with larger ranges from dominating the model's learning process \cite{marins2021fault}. Both techniques ensure the model's sensitivity to minority class features.

\textbf{\textit{IV) Dimensionality reduction}}:
Dimensionality reduction is more than a standard preprocessing step; it is essential for rare event prediction as it addresses the high dimensionality and complexity inherent in rare event datasets. Principal Component Analysis (PCA) reduces the feature space by preserving the most informative components that distinguish rare events from normal occurrences \cite{xu2021two, rehab2021bearings, bhanja2022black, abbasi2022large, fathy2020learning, marins2021fault, omar2022exploring, ravindranath2020m2nn, alestra2014rare, xu2022training, abbasi2022large}. Focusing on the principal components, PCA filters out noise and highlights subtle signals, which aids models in learning from imbalanced data.

\textbf{\textit{V) Fourier transform}}:
This is widely used in signal processing and image processing. It transforms a time-domain signal into its frequency-domain representation, allowing for analyzing and manipulating the signal's frequency components. In rare event prediction, it can be used to uncover frequency-domain characteristics that reveal rare occurrences often hidden by time-domain noise. In audio-based rare event research, \cite{abbasi2022large}, fourier transform is applied to convert audio waveforms into the frequency domain for further analysis.

\subsubsection{Analyzing feature engineering approaches with data modalities, rarity groups, and downstream tasks}
\label{sec:fe1}

\begin{figure}[h]
  \centering
  \includegraphics[width=0.8\linewidth]{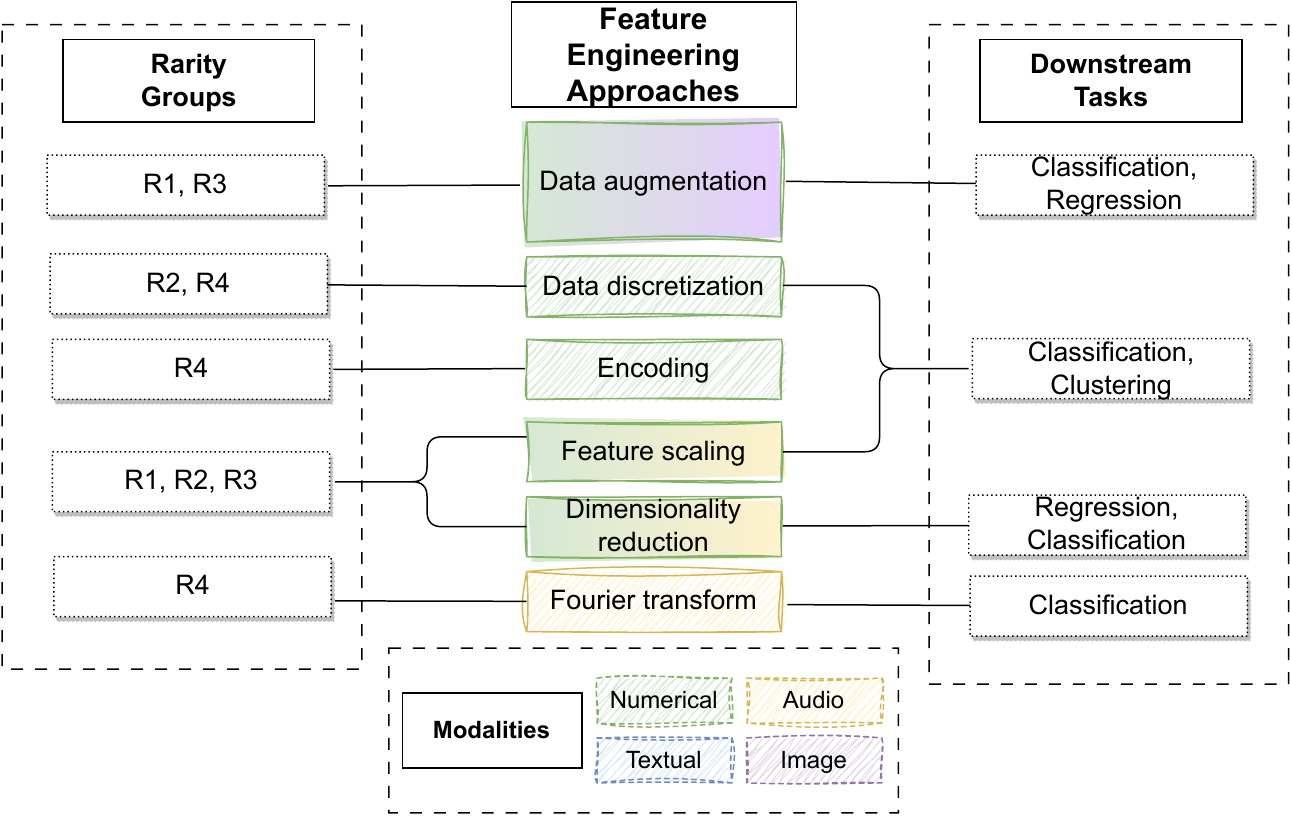}
  \caption{Association between feature engineering approaches, data modalities, rarity groups and downstream tasks \\
  * Coloring of feature engineering approaches corresponds to the data modalities
}
  \label{fig:featureengineerning}
  
\end{figure}

Figure \ref{fig:featureengineerning} presents a comprehensive overview of the association between feature engineering techniques, data modality, and rarity groups in the context of rare event prediction, as examined from the reviewed papers.  It is observed that classification has been the primary focus of the majority of research. Standardization, normalization, and dimensionality reduction have been applied to numerical and audio data, whereas data augmentation has been used on numerical and image data. Discretization and encoding have been used with numerical data, and in addition to classification, these techniques focus on clustering tasks.  It is noted that none of the feature engineering techniques are rarity-independent; hence, each of the techniques seems to perform well with specific rarity groups.

\begin{table}[!ht]
\footnotesize	
\caption{Data processing approaches vs. rarity groups, modality, and downstream tasks. $^*$} 
\label{tab:dataprocessing}
\begin{tabular}{llllll}
\toprule
\textbf{Data processing approach} &
  \textbf{Papers} &
  \textbf{Rarity group} &
  \textbf{\begin{tabular}[c]{@{}l@{}}Downstream \\ tasks\end{tabular}} &
  \textbf{Modality} &
  \textbf{\begin{tabular}[c]{@{}l@{}} Dataset Type\end{tabular}} \\
  \midrule

\multicolumn{6}{l}{\textbf{1. Data Cleaning}} \\
Data sifting &
  \cite{cheon2009bayesian, meng2006rare, gujrati2007meta, lan2010study} &
  R1, R2, R3, R4 &
  FT, CL, CF  &
  N &
  RE, DE \\
Data filtering &
  \cite{hsieh2019unsupervised, xu2022new, xu2022training, meng2006rare} &
  R1, R2, R3, R4 &
  CF &
  N &
  RE \\
Imputation &
\begin{tabular}[c]{@{}l@{}}
  \cite{ravindranath2020m2nn, seiffert2007mining, ahmadzadeh2019rare, cheon2009bayesian, adil2022deep}, \\ \cite{omar2022exploring, gondek2016prediction, rafsunjani2019empirical, xiu2021variational, omar2022exploring, rafsunjani2019empirical} {}
  \end{tabular}&
  R1, R2, R3, R4 &
  \begin{tabular}[c]{@{}l@{}}
  CL, CF, FT 
  \end{tabular}&
  N &
  \begin{tabular}[c]{@{}l@{}}RE, DE, SIY \end{tabular} \\
Noise removal &
  \cite{omar2022exploring, li2017rare, iyer2015statistical, ashraf2023identification}{]} &
  R2, R3, R4 &
  CL, CF &
  N &
  RE, DE \\
Textual summary generation &
  \cite{wong2016statistical, zhao2018framework} &
  R4 &
  CF &
  TX &
  RE \\
Text conversions &
  \cite{wong2016statistical} &
  R4 &
  CF &
  TX &
  RE \\
Stemming and lemmatization &
  \cite{wong2016statistical} &
  R4 &
  CF &
  TX &
  RE \\
Image processing techniques &
  \cite{kubat1998machine} &
  R2 &
  CF &
  I &
  RE \\
  \cline{1-6} \\
\multicolumn{6}{l}{\textbf{2. Feature Selection}} \\
\begin{tabular}[c]{@{}l@{}} Correlation-based  \\ feature selection\end{tabular} &
  \cite{li2017rare, dangut2022application, olmucs2022comparison} &
  R1, R1, R3 &
  CF, SM &
  N &
  \begin{tabular}[c]{@{}l@{}}RE, DE, SIY \end{tabular} \\
\begin{tabular}[c]{@{}l@{}}Wrapper-based methods \end{tabular} &
  \cite{rehab2021bearings} &
  R4 &
  CL &
  N &
  RE \\
\begin{tabular}[c]{@{}l@{}}Statistical measures\end{tabular} &
  \begin{tabular}[c]{@{}l@{}}\cite{alestra2014rare, ahmadzadeh2019rare, gondek2016prediction, bhanja2022black, marins2021fault, xu2021two}, \\ \cite{wong2016statistical, zhao2018framework}\end{tabular} &
  R1, R2, R4 &
  CF, FT &
  N, TX &
  RE, SIY \\
\begin{tabular}[c]{@{}l@{}} Gini feature importance and \\ XGBoost accuracy and \\ cover measurement\end{tabular} &
  \cite{chen2004using, omar2022exploring, hebert2016predicting, gondek2016prediction, omar2022exploring} &
  R1, R2, R3, R4 &
  CL, CF &
  N &
  RE \\
\begin{tabular}[c]{@{}l@{}} MFCC-based feature extraction\end{tabular} &
  \cite{abbasi2022large, shi2020few} &
  R4 &
  CF &
  A &
  RE \\
\begin{tabular}[c]{@{}l@{}}Intrinsic-based feature selection\end{tabular} &
  \begin{tabular}[c]{@{}l@{}}\cite{ravindranath2020m2nn, li2017rare, kubat1998machine}\end{tabular} &
  R1, R2, R3, R4 &
  CF, RG &
  N, I &
  RE, DE \\
  \cline{1-6} \\
\multicolumn{6}{l}{\textbf{3. Sampling}} \\
\begin{tabular}[c]{@{}l@{}}\textit{Basic sampling}\\ ~~~i) ROS \& RUS\end{tabular} &
  \begin{tabular}[c]{@{}l@{}}\cite{zhao2018framework,nugraha2020clustering, wu2007local, wong2016statistical, ranjan2018dataset}, \\ \cite{wang2012applying, hamaguchi2019rare, ahmadzadeh2019rare, iyer2015statistical, nugraha2020clustering, chen2004using}\end{tabular} &
  R1, R2, R3, R4 &
  CF, FT, CL &
  N, I, TX &
  RE, DE \\
\textit{Advanced sampling} &
   &
   &
   &
   &
   \\
~~~i) SMOTE &
  \cite{zhao2018framework, chen2004using, wong2016statistical, li2017adaptive, ali2014dynamic, dangut2022application, seiffert2007mining} &
  R1, R2, R3, R4 &
  CF &
  N, TX &
  RE, DE \\
~~~ii) ADASYN &
  \cite{ashraf2023identification} &
  R3 &
  CF &
  N &
  RE, DE \\
~~~iii) SMUTE &
  \cite{li2017rare} &
  R3, R4 &
  CF &
  N &
  \begin{tabular}[c]{@{}l@{}}RE, DE, SIY \end{tabular} \\
~~~iv) ENN &
  \cite{tomek1976experiment, li2017rare, ashraf2023identification} &
  R3, R4 &
  CF &
  N &
  \begin{tabular}[c]{@{}l@{}}RE, DE, SIY \end{tabular} \\
~~~v) NCL &
  \cite{li2017rare, ashraf2023identification} &
  R3, R4 &
  CF &
  N &
  \begin{tabular}[c]{@{}l@{}}RE, DE, SIY \end{tabular} \\
~~~vi) NearMiss &
  \cite{li2017rare, ashraf2023identification} &
  R3, R4 &
  CF &
  N &
  \begin{tabular}[c]{@{}l@{}}RE, DE, SIY \end{tabular} \\
~~~vii) OSS &
  \cite{kubat1997addressing, kubat1998machine, chen2004using, seiffert2007mining} &
  R2, R3 &
  CF &
  N, I &
  RE, DE \\
~~~viii) CBO &
  \cite{jo2004class, seiffert2007mining} &
  R2, R3, R4 &
  CL, CF &
  N &
  DE \\
~~~ix) Time series subsampling &
  \begin{tabular}[c]{@{}l@{}}
  {\cite{fukuchi1999subsampling, combes2022time, liu2021machinery}}
  \end{tabular} &
  R3, R4 &
  CF &
  N &
  RE \\
~~~x) Stratified sampling &
  \cite{liu2021machinery} &
  R4 &
  CF &
  N &
  RE \\
~~~xi) Data framing &
  \cite{abbasi2022large} &
  R4 &
  CF &
  A &
  RE \\
~~~xii) Uncertainty sampling &
  \cite{pickering2022discovering} &
  R1 &
  FT, SM &
  N &
  SIY \\
\begin{tabular}[c]{@{}l@{}}
~~~xiii) Choice-based \slash \\ ~~~~~~~~~~~~endogenous sampling\end{tabular} &
  \cite{maalouf2018logistic, van2006prediction} &
  R1,R2,R3, R4 &
  CF &
  N, I &
  RE \\
\cline{1-6} \\
\multicolumn{6}{l}{\textbf{4. Feature Engineering}} \\
Data augmentation &
  \cite{fathy2020learning, ashraf2023identification, Ranjan2019DataCD, ranjan2018dataset, yang2022applications, hamaguchi2019rare, parsa2021deep} &
  R1, R3 &
  \begin{tabular}[c]{@{}l@{}}
  CF, RG, CL 
  \end{tabular}&
  N, I &
  RE \\
Data discretization &
  \cite{omar2022exploring, seiffert2007mining, neuman2021extreme, berberidis2007detection} &
  R2, R4 &
  CL, CF &
  N &
  RE, DE \\
Encoding &
  \cite{liu2021machinery} &
  R4 &
  CL, CF &
  N &
  RE \\
Feature scaling &
  \begin{tabular}[c]{@{}l@{}}\cite{liu2021machinery, lee2021early, xu2022training,  abbasi2022large, marins2021fault, ranjan2020understanding, martello2021improving}\end{tabular} &
  R1, R2, R4 &
  CL, CF &
  N, A &
  RE, DE \\
Dimensionality reduction &
\begin{tabular}[c]{@{}l@{}}
  \cite{xu2021two, rehab2021bearings, bhanja2022black, abbasi2022large, fathy2020learning, marins2021fault}, \\ \cite{omar2022exploring, ravindranath2020m2nn, alestra2014rare, xu2022training, abbasi2022large} \end{tabular} &
  R1, R2, R4 &
  \begin{tabular}[c]{@{}l@{}}
  RG, CF, FT 
  \end{tabular}&
  N,A &
  RE, DE \\
Fourier transform &
  \cite{abbasi2022large} &
  R4 &
  CL &
  A &
  RE \\
  \bottomrule
\end{tabular}
\\\footnotesize{$^*$ N-Numeric, TX-Textual, I-Image, A-Audio, T-Time series, FT-Forecasting, CL-Clustering, CF-Classification, RG-Regression, RE-Naturally rare, DE-Derived, SIY-Simulated/Synthetic}

\end{table}

In this section, we explored four primary techniques, data cleaning, feature selection, sampling, and feature engineering, as approaches to data processing. Data cleaning methods help to ensure data quality and remove noise, while feature selection techniques aim to identify the most relevant features for rare event prediction. Sampling approaches assist in selecting a subset of datasets and address issues with class imbalance and large dataset sizes. Feature engineering methods effectively extract meaningful information and represent the data more discriminatively. 
We also examined their application in various data modalities, rarity levels, and downstream tasks. It is observed that many studies utilize standard data processing techniques common to general ML research. This highlights the need to investigate standardized approaches adapted to rare events to assist the unique challenges of classification, clustering, regression, and forecasting. Notably, numerical data cleaning techniques like data sifting, filtering, imputation, and noise removal are commonly used, while feature selection methods such as correlation-based, wrapper-based, and filter-based techniques are applied across different rarity groups, with classification tasks being the primary focus across all methodologies.

\section{Algorithmic Approaches}

Algorithmic approaches are pivotal in any ML pipeline and contribute significantly in making informed and effective decisions. They provide mathematical models for varied use cases centered around downstream tasks like classification, clustering, forecasting, regression, and simulation. Firstly, this section analyzes a subset of algorithmic approaches utilized in the literature on rare events. Then, each approach would be examined concerning several algorithmic indicators. Finally, each approach would be analyzed with the data modality, rarity groups, data processing techniques, and downstream tasks.

\begin{figure}[!ht]
  \centering
  \includegraphics[width=0.8\linewidth]{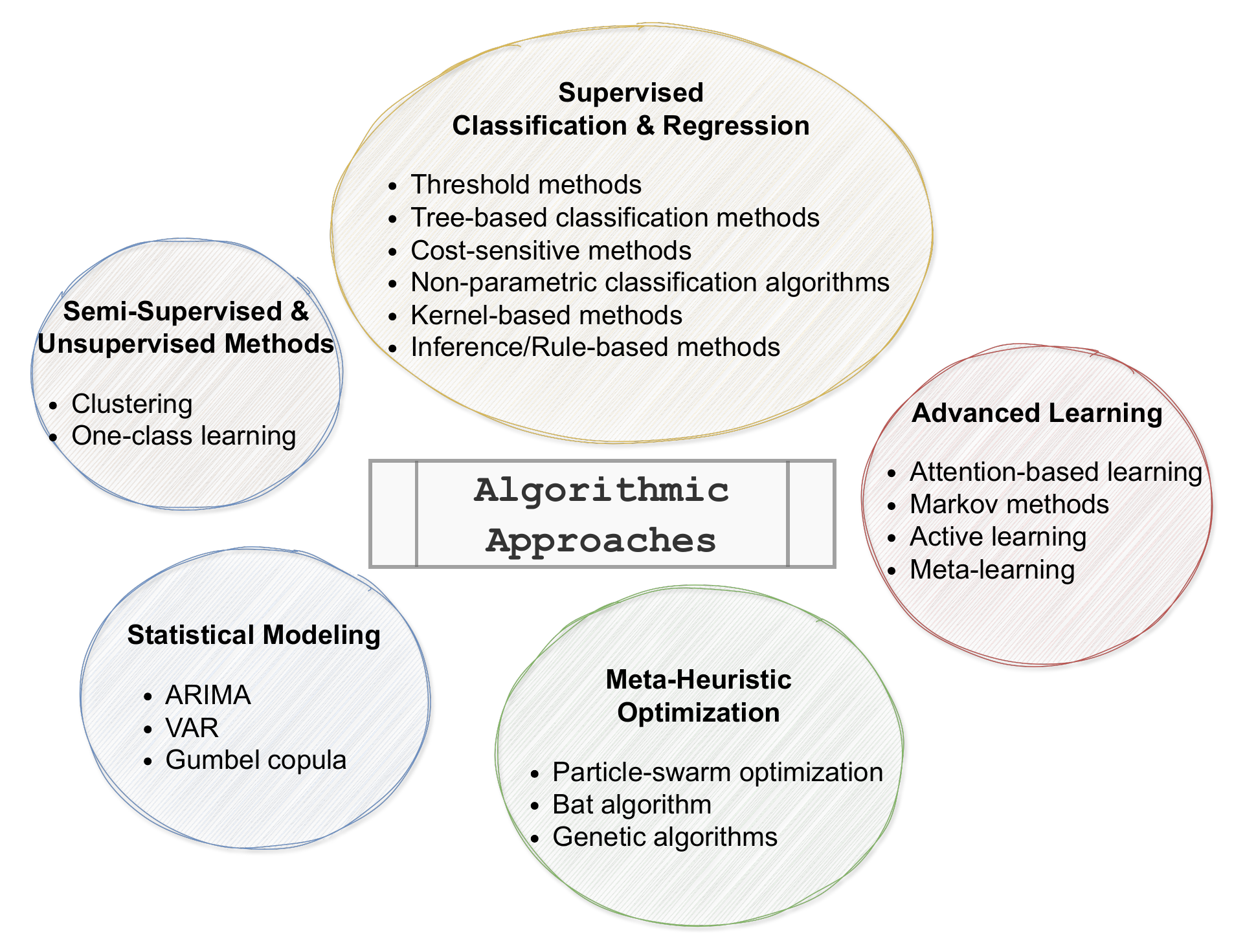}
  \caption{Algorithmic approaches used in rare-event prediction}
  \label{fig:algorithms}
\end{figure}

\subsection{Significance of algorithmic approaches}
Algorithmic approaches provide tools and modeling techniques to analyze and interpret complex datasets, enabling the identification of patterns, relationships, and key factors and predicting events of significant importance. We categorize algorithmic approaches in rare event research into five major groups: Supervised classification and regression, Clustering, Statistical modeling, Meta-heuristic optimization, and Advanced learning methods as shown in Figure \ref{fig:algorithms}.

\subsection{Supervised Classification and Regression Methods}
Classification and regression are two fundamental supervised learning tasks in machine learning aimed at predicting the output of a target variable based on input features. The key difference lies in the type of target variable they handle. Classification algorithms are used to predict discrete values such as gender, binary labels (true/false), or categories like spam or not spam, while regression algorithms are employed for predicting continuous values like price, salary, or age. In rare event prediction, various techniques have been used, including threshold methods, tree-based classification, one-class learning, cost-sensitive methods, non-parametric classification algorithms, kernel-based methods, and inference/rule-based methods. 

\subsubsection{Threshold methods}
These approaches set a specific threshold for classifying data instances into rare or non-rare classes based on a pre-defined criterion, rendering them well-suited for modeling rare events. Instances above the threshold are considered rare, while those below it are classified as non-rare. Existing research has employed probabilistic statistical methods such as Logistic Regression (LR), Naive Bayes classifiers (NB), and Neural Networks (NN) that generate a score or probability threshold. If the categorization is binary, this probability is subsequently transferred and mapped to a binary mapping, like malignant or benign, spam or not spam, normal or abnormal. 
In rare event prediction, certain researchers have utilized LR as a fundamental base classifier \cite{zhao2018framework, van2006prediction}, while others have explored various adaptations and variations, including incorporating regularization techniques \cite{alestra2014rare, ashraf2023identification} and utilizing weighting methods \cite{maalouf2018logistic}. It was observed that the conventional LR encountered convergence challenges in scenarios with limited sample sizes or rare events in the data. To address this issue, an alternative approach known as Firth's logistic regression \cite{puhr2017firth} has been employed, wherein a penalty is added to the log-likelihood function of the LR model. Several studies that include rare events \cite{olmucs2022comparison, ashraf2023identification} have demonstrated improved performance using Firth's logistic regression. LR methods are easy to implement, interpretable, and suitable for rare event prediction with small sample sizes.

Bayes classifiers and neural networks are also threshold-based models utilized in rare event prediction research. Bayes classifiers have been used in rare event prediction by \cite{rafsunjani2019empirical, li2017rare}. Studies have explored the application of neural networks, including Deep neural networks \cite{adil2022deep, bai2022rare, li2017rare}, Convolutional Neural Networks (CNN) \cite{bai2022rare, bhanja2022black, yang2022applications, parsa2021deep, lee2021early}, Multi-Layer Perceptron (MLP)  \cite{abbasi2022large}, Autoencoders \cite{ ashraf2023identification, martello2021improving, xu2022training}, CNN-based autoencoders \cite{bhanja2022black, ranjan2020understanding, martello2021improving, xu2022training, hsieh2019unsupervised}, Long Short-Term Memory (LSTM) autoencoders \cite{ranjan2020understanding, hsieh2019unsupervised} in various applications such as mineral prospectivity prediction \cite{yang2022applications, parsa2021deep}, black-swan event prediction \cite{bhanja2022black}, rare sound classification in audio forensics \cite{abbasi2022large} and manufacturing failure prediction \cite{ranjan2020understanding, hsieh2019unsupervised}. LSTMs can preserve temporal properties of rare events by capturing sequential dependencies and managing long-term relationships through gating mechanisms. NB offers robustness, probabilistic inference, and flexibility compared to LR, while NN excels in representation learning and scalability. But, both risks being affected by the insufficient sample size in rare events \cite{olmucs2022comparison}.

\subsubsection{Tree-based classification methods}
These are supervised ML methods that partition the training data into subsets using a series of conditional statements. These splits create a tree-like structure, each representing a logical test on a specific feature. The final model comprises multiple such trees, enabling predictions and offering insights into the relationships within the data. Tree-based classification methods are widely used in rare event prediction due to their ability to handle complex and non-linear relationships in the data. Random Forest (RF) \cite{coley2023empirical,bai2022rare, gondek2016prediction, fathy2020learning,hebert2016predicting, rafsunjani2019empirical, marins2021fault} and Boosted Classification Trees, such as XGBoost \cite{hebert2016predicting, fathy2020learning, ashraf2023identification, rafsunjani2019empirical}, are popular choices in this group. In rare event prediction, researchers have applied these models to estimate the probability of rare events like suicide attempts \cite{coley2023empirical}, APS failures \cite{gondek2016prediction, rafsunjani2019empirical}, and manufacturing faults \cite{fathy2020learning, hebert2016predicting}. Data augmentation techniques have been used to enhance the performance of tree-based models further, optimizing parameters to avoid overfitting \cite{ashraf2023identification}. Tree-based methods also benefit from handling overfitting using random subsets during model training. However, they can be challenging to interpret due to the large number of trees in the model, making it difficult to understand the combined effect of all trees. Despite this limitation, tree-based classification methods are valuable tools for capturing complex relationships and making accurate predictions in rare event prediction tasks.

\subsubsection{Cost-sensitive learning}
These methods consider the costs associated with prediction errors and other potential costs during the training of a ML model. Instead of maximizing accuracy, the focus shifts to minimizing overall misclassification costs, where each class or instance is assigned a specific misclassification cost. False negatives (misses) are assigned higher costs than false positives (false alarms). Two main approaches to cost-sensitive learning are decision trees and weighting. Decision trees employ a parameterized threshold mechanism to dynamically adjust the decision boundary of the classifier, making them suitable for modeling rare events in a manner that is nuanced and contextually adaptable \cite{zhao2018framework, li2017rare, kubat1998machine, nugraha2020clustering}. On the other hand, weighting assigns higher weights to the minority class to penalize misclassifications of the rare class \cite{he2021weighting, chen2004using}. Researchers have explored cost-sensitive learning methods with various classifiers such as Logistic Regression, Random Forest, and Support Vector Machines. Weighted Random Forest and AdaClassWeight are examples of algorithms that adaptively adjust the weights of the rare class to address the imbalanced data problem \cite{he2021weighting, chen2004using}. Overall, it's observed that cost-sensitive learning methods offer effective solutions for handling imbalanced datasets and predicting rare events in various real-world applications.

\subsubsection{Non-parametric classification algorithms}
They are ML algorithms that do not make explicit assumptions about the data's underlying probability distribution or functional form. Unlike parametric classification algorithms, which assume a specific functional form (linear or polynomial) and estimate parameters, non-parametric algorithms learn directly from the data without assuming any specific model structure. K-nearest neighbors (k-NN) is an instance adopted by \cite{rafsunjani2019empirical} in APS failure prediction. They are often more flexible and can capture complex relationships in the data, which becomes more advantageous in rare event prediction \cite{lee2021early, rafsunjani2019empirical}. 

\subsubsection{Kernal-Based Methods}
They are ML algorithms that transform data into a higher-dimensional space using kernel functions, enabling them to capture non-linear relationships and solve downstream tasks like classification and regression. Support Vector Machines (SVMs) are a well-known kernel technique that outperforms neural networks in some rare event research, particularly for small to medium datasets demanding explainable outcomes. SVM-based kernels can disambiguate hard-to-classify rare event datasets using soft-margins \cite{iyer2015statistical,ahmadzadeh2019rare, rafsunjani2019empirical}. One-class SVM has been used in evaluating large sample sizes in unsupervised learning environments \cite{alestra2014rare, iyer2015statistical}. In imbalanced healthcare data, L.SVM (Support Vector Machine with Linear kernel) and R.SVM (Support Vector Machine with Radial kernel) have been used as base classifiers for incident detection \cite{zhao2018framework}. SVMs have also been applied in importance sampling for rare event detection to identify multiple failure regions and estimate the structural failure probability of rare events \cite{ling2021support}. Granular SVM (GSVM) is a variation of SVM that combines statistical learning theory and granular computing theory \cite{tang2004granular}. In \cite{tang2008svms}, GSVM has been used with under-sampling techniques to improve efficacy by locally reducing redundant data through parallel processing. Rare event weighted kernel logistic regression (RE-WKLR) \cite{maalouf2011robust} is an algorithmic SVM enhancement optimized for unbalanced and rare event data. It offers the advantages of weighing, bias correction, and the strength of kernel approaches, especially when datasets are imbalanced or not linearly separable.

\subsubsection{Inference/Rule-Based Methods}
This class of algorithms focuses on deriving knowledge and insights from data through explicit if-then rules. 
In rare event literature, these methods can be Bayesian methods, Inductive algorithms, Two-phase rule induction, Knowledge-based and human interaction-based approaches, and Association rule mining. These methods are applicable to rare event prediction due to their ability to leverage expert domain knowledge, create human-readable rules, and provide interpretable outputs, which are crucial for understanding the decision-making process in scenarios where data scarcity and complexity make traditional modeling approaches challenging.

\textbf{\textit{i) Bayesian methods}}: 
These are statistical techniques that use prior information on a certain population, and they rely on Bayes' theorem to update the probability of a hypothesis based on new evidence. Bayesian methods offer multiple benefits, including handling incomplete and noisy data, understanding causal relationships between variables, and integrating domain expertise and data through Bayesian networks. Utilizing Bayesian methods, as demonstrated in \cite{cheon2009bayesian}, along with Bayesian networks, results in superior performance in forecasting daily ozone states by incorporating expert knowledge and historical data. Bayesian networks also prove useful for extreme rare event identification, as shown in \cite{neuman2021extreme}, where the Jaynes inferential approach is used for feature engineering and optimal binning, leading to reduced features and improved identification of relevant diagnostic features.

\textbf{\textit{ii) Inductive algorithms}}:
In rare event prediction, some researchers have proposed various inductive bias methods such as Maximum Specificity Bias (MSB) and Instance-Based Learning (IBL). MSB aims to discover specific rules for individual training examples, enhancing the performance of small disjuncts but leading to worse overall performance.  IBL, also known as lazy learning, is a ML approach that focuses on the local generalization of instances based on similarity measures. 1-NN (1-Nearest Neighbor) algorithm is a type of IBL used in \cite{kubat1998machine} along with one-sided selection in detecting oil spills. This study shows promising results for improving accuracy in small disjuncts, but even this method has not been able to provide conclusive evidence. 

\textbf{\textit{iii) Two-phase rule induction}}:
Is a ML technique that involves a two-step process for inducing rules from data and is commonly used in data mining and knowledge discovery. The PNrule algorithm \cite{agarwal2001pnrule} is a two-phase rule induction approach that involves the discovery of positive rules (P-rules) to predict the presence of a class and negative rules (N-rules) to predict its absence. P-rules are learned in the first phase to capture the most positive cases while maintaining a respectable false positive rate. N-rules are discovered in the second phase to reduce false positives introduced by the union of P-rules while maintaining an acceptable detection rate. The study by \cite{joshi2001mining} uses the PNRule method to address the challenge of maximizing recall and precision in rare event prediction. By creating rules with great accuracy, the initial phase concentrates on recall. The second phase concentrates on precision by utilizing rules that eliminate false positives from the records covered by the first phase.

\textbf{\textit{iv) Knowledge-based and human interaction-based approaches}}:
Knowledge and human interactions have been researched and explored in predicting rare events like international conflicts, wars, coups, revolutions, massive economic depressions, and economic crises \cite{king2001explaining}. Knowledge-driven models are particularly useful in areas with limited exploration data or that have not been extensively studied. They rely on the expertise of professionals to make decisions, but they can be subjected to limitations due to their subjective nature. Some instances of using knowledge include modeling knowledge from geological experts in mineral prospectivity prediction \cite{yang2022applications, parsa2021deep} and using expert knowledge like chemical equations and hypotheses in solar flare forecasting \cite{cheon2009bayesian}. Nevertheless, while these approaches may demonstrate efficacy when applied to small and straightforward systems, their effectiveness might be low when dealing with more complex and diverse systems \cite{dangut2022application}.

\textbf{\textit{v) Association rule mining}}:
These utilize basic If/Then statements to reveal relationships between independent relational or other data repositories. \cite{vilalta2002predicting} adopts an association rule mining approach to identify frequently occurring patterns preceding target rare events, which are subsequently integrated into a rule-based predictive model.  `PREVENT' is a general purpose inter-transaction association rules mining algorithm that uses inter-transactional patterns to predict rare events in transactional databases \cite{berberidis2007detection}. FP-Growth, a state-of-the-art algorithm for classical association rule mining, is utilized there. The algorithm's computational cost is minimal as it involves limited scans of the database, making it well-suited for the requirements of rare event prediction. 

\subsection{Semi-supervised and Unsupervised methods}
Semi-supervised methods for rare event prediction leverage a combination of labeled data from the rare event class and unlabeled data from normal instances, aiming to improve predictive accuracy and generalization. In contrast, unsupervised methods rely solely on unlabeled data, focusing on identifying patterns or anomalies that deviate significantly from the norm, which can be indicative of rare events.

\subsubsection{Clustering methods}
Clustering is a type of unsupervised learning where samples are categorized based on their resemblance to neighboring data points. Several clustering-based methods have been employed in the literature on rare events. Distance-based unsupervised methods such as Random Forest (RF) clustering, Partition Around Medoids (PAM), K-means, and hierarchical clustering are commonly used \cite{lazarevic2004data, xu2022training}. For instance, hierarchical clustering has been used to detect abnormal production behaviors in the paper manufacturing industry \cite{xu2022training}. K-means clustering has been utilized to identify rare change events based on the distance between common features \cite{hamaguchi2019rare}. Additionally, clustering-based undersampling techniques, like Clustering Large Applications (CLARA)  \cite{nugraha2020clustering} and Classification using lOcal clusterinG (COG) algorithms \cite{wu2007local}, have aimed to generate balanced sub-classes for classification. In some cases, ensembles of clustering methods, such as RF Clustering and PAM, are been combined to optimize the clustering process  \cite{omar2022exploring}. Moreover, the combination of nearest neighbor and Balanced Iterative Reducing and Clustering using Hierarchies (BIRCH) clustering with dynamic markov chains has shown promise in detecting rare events in spatiotemporal environments using sensor and traffic data  \cite{meng2006rare}.

\subsubsection{One-class learning}
One-class learning is an unsupervised learning strategy used for extremely skewed class distributions, with the classifier being trained exclusively on data from one class. It can be considered a semi-supervised or unsupervised approach, depending on how it is implemented and the specific context of its application.
Adaptation of one-class classification algorithms for imbalanced classification has been researched in early studies, and they have also been employed in rare event prediction. HIPPO (i.e.,a classification method based on Hippocampus functioning) \cite{japkowicz1995novelty} is a standard method where only the rare class is learned, and Repeated Incremental Pruning to Produce Error Reduction (RIPPER) \cite{cohen1995fast} is a standard method where the algorithm selects the majority class as its default class and learns the rules for detecting the minority class. Hamaguchi et al. \cite{hamaguchi2019rare} have proposed a variational autoencoder-based method to learn disentangled representations on only low-cost negative samples of image data. As a result, rare events were detected as outliers. Some autoencoder models we presented under threshold methods also do intersect with this group, such as \cite{ranjan2020understanding}, which was trained only on normal samples, and \cite{ashraf2023identification}  where samples of rare events were used for training. 
The advantages of one-class classifiers come at the cost of ignoring all available information about one class; consequently, this solution should be approached cautiously, as it may not be suitable for all circumstances.

\subsection{Statistical Modeling}
Statistical models are algorithms that apply statistics and mathematics concepts to generate a representation of data, which is then analyzed to determine any relationships or discover insights. Statistical approaches like Autoregressive Integrated Moving Average (ARIMA) modeling and Vector Autoregression (VAR) have been used in rare event prediction \cite{alestra2014rare, hsieh2019unsupervised}. ARIMA is an autoregressive statistical model that predicts future values based on past values. Alestra et al. have utilized ARIMA modeling separately for each aircraft to predict the degradation behavior over time \cite{alestra2014rare}. Vector Autoregression follows a stochastic process model that captures linear interdependencies among multiple time series using a linear function of past values of each variable. \cite{hsieh2019unsupervised} explored VAR in evaluating the LSTM-Autoencoder model developed. ARIMA and VAR models effectively capture time-based dependencies and sequential (temporal) data patterns, which enhances prediction accuracy and addresses challenges in predicting rare events based on past observations. The Gumbel copula function is a statistical modeling technique in copula-based algorithms utilized in rare event research. Copulas are mathematical functions used to model the dependence structure between random variables. They describe the joint distribution of variables by connecting their marginal distributions. The Gumbel copula function, specifically, is a type of copula that is based on the Gumbel distribution. It is commonly used to model extreme-value dependence, making it suitable for applications involving rare events or extremes. It captures the tail dependence between variables, which is vital in scenarios where the behavior of variables in the extreme tails is of interest \cite{yang2022prediction}.

\subsection{Meta-Heuristic Optimization}
Meta-heuristic algorithms are optimization techniques for solving complex problems by iteratively exploring and exploiting the search space to find near-optimal solutions \cite{abdel2018metaheuristic}. These algorithms, such as particle swarm optimization and the bat algorithm, have been applied to improve efficiency and scalability on large imbalanced healthcare data, \cite{li2017adaptive}, and genetic algorithms have been utilized for learning to predict rare events in categorical time series data \cite{ weiss1998learning}. Additionally, evolutionary ensemble algorithms have shown promise in enhancing the identification rate of minority classes in rare event-based imbalanced datasets \cite{krawczyk2014cost}. These algorithms would excel in rare event prediction due to their ability to efficiently explore complex search spaces and adaptively exploit the underlying patterns and structures in imbalanced datasets.

\subsection{Advanced Learning Methods}
Recently, advanced learning methods like Attention-based mechanisms, Markov methods, Active learning, and Meta-learning have emerged to tackle the challenges posed by rare event datasets. These approaches strive to enhance rare event predictive capability by leveraging temporal dependencies, probabilistic modeling, and selective data labeling.

\subsubsection{Attention-based mechanisms}
Attention mechanisms enable the model to selectively concentrate on important input elements for accurate predictions while disregarding less relevant parts. This has gained attention in rare event research as well. For instance, Liu et al. proposed a fault diagnosis approach using one-dimensional CNN, Gated Recurrent Unit (GRU), and attention mechanisms, combined with knowledge graphs, to achieve more precise predictions in bearing fault detection \cite{liu2021machinery}. Xu et al. have utilized attention-based-LSTM and Extra-Tree models for fault mode and severity prediction based on bearing datasets, while Ravindranath et al. introduced the M2NN architecture with an attention mechanism for post-traumatic seizure detection, demonstrating its effectiveness in finding unusual seizures in EEG data \cite{xu2021two, ravindranath2020m2nn}. Kulkarni and team utilized soft-attention CNN for rare marker event detection and localization in well-logs \cite{kulkarni2020soft}. Attention-based mechanisms preserve temporal properties by selectively focusing on important time steps, capturing both short- and long-term dependencies in sequential data. These methods have proven useful in handling rare event detection tasks, such as anomaly detection in manufacturing, rare disease diagnosis, and identifying specific rare occurrences in multiple domains.

\subsubsection{Markov methods}
Markov models represent the Markov property, which asserts that the prediction of an outcome depends only on information about the current state and is independent of the previous sequence of events. Extensible Markov Models (EMM) and Monte Carlo methods are among the markov methods adopted in rare event literature. These models capture the time-dependent behavior by modeling transitions between states over time and, hence, can address the temporal properties inherent in rare events.

\textbf{\textit{I) Extensible markov models (EMM)}}:
Extensible Markov Models (EMM) are a dynamic variation of traditional static Markov models. They excel in modeling spatiotemporal data and have proven to help predict spatiotemporal events, including rare occurrences, by capturing spatial, temporal, and unusual event transitions \cite{dunham2004extensible}. Meng et al. \cite{meng2006rare} introduced an approach that combines clustering and EMM for rare event detection in spatiotemporal data, particularly sensor and traffic data. Their contribution lies in modeling the transitions between rare and typical events using EMM, effectively detecting rare occurrences.

\textbf{\textit{II) Monte carlo methods}}:
Monte carlo methods, also called Monte carlo simulations, are probabilistic mathematical techniques that estimate the possible outcomes of uncertain events. In rare event estimation, Monte Carlo methods have been applied to estimate model uncertainty, such as using Monte Carlo Dropout to create multiple predictions and measuring the standard deviation of detection depths as a proxy for uncertainty \cite{kulkarni2020soft}. Additionally, they have been used to predict rare events using trajectory data and unscheduled aircraft maintenance actions \cite{dang2022parallel, githubChaoDangRareEventEstimation, olmucs2022comparison, bai2022rare, ling2021support, wen2020batch, strahan2023inexact, dangut2022application}.

\subsubsection{Active learning}
Active learning is a learning strategy that iteratively chooses the most instructive or uncertain instances from a pool of unlabeled data and asks an oracle (such as a human expert or a pre-existing labeled dataset) to annotate their labels \cite{settles2009active}. Recent research efforts have been to predict rare events using active learning, exemplified by studies like \cite{dhulipala2022active, pickering2022discovering}. Dhulipala et al. \cite{dhulipala2022active} proposed a methodology for rare event simulation by combining active learning and multi-fidelity modeling, which leverages multiple levels of models' fidelity (accuracy or complexity) to predict outcomes efficiently. Deep Neural Operators (DNOs), like DeepONet \cite{lu2021learning}, are nonlinear operators created to handle systems with infinite dimensions, making them highly effective surrogate models for precisely representing extreme events. Pickering et al. \cite{pickering2022discovering} introduced a Bayesian-inspired framework based on active learning for DNOs, specifically designed for discovering and quantifying extreme events, focusing on uncertainty quantification.

\subsubsection{Meta learning}
Meta-learning, or "learning to learn", is a ML approach focusing on training models to effectively adapt and generalize to new tasks or domains with minimal data. It can be applied to rare event prediction by training a model on various rare event scenarios. With limited data, it can rapidly adapt and accurately predict new and infrequent events. The advantage of meta-learning over other advanced learning approaches lies in its ability to dynamically adapt and integrate different predictive techniques,  a faster and cheaper training process, thus enhancing the model's generalization capability and performance across varied scenarios. Several studies \cite{lan2010study, gujrati2007meta, shi2020few} have looked at how meta-learning techniques can be used to improve failure prediction in different areas, such as large-scale computing systems and acoustic event detection, by dynamically combining base methods and improving accuracy in situations with few labeled data. While this existing research primarily focuses on failure prediction, it is worth noting that no prior studies have specifically addressed the prediction of rare events using meta-learning techniques.

\subsection{Comparison of algorithmic approaches}
Table \ref{tab:algorithms1} offers a comparative analysis of the approaches discussed in the preceding section. We use 12 algorithmic indicators based on four major factors: computational efficiency, model analysis and understanding, data availability and model performance. Table \ref{tab:algorithms2} analyses algorithmic approaches with rarity groups, downstream tasks, modalities, dataset types, and data processing tasks based on our reviewed papers. Predominantly, research has centered on classifying numeric rarity within real datasets, showcasing a blend of various data processing methodologies. Limited research is dedicated to alternative downstream tasks such as clustering, forecasting, simulation, and diverse modalities encompassing text, images and audio.

\begin{table}[!ht]
\scriptsize
\caption{Comparison of algorithmic approaches}
\label{tab:algorithms1}
\begin{tabular}{clccccc}
\toprule
\textbf{Major Factor}                                                                                     & \textbf{Algorithmic Indicator}                                                                  & \textbf{\begin{tabular}[c]{@{}l@{}}Supervised\\ Classif. \& Regress.\end{tabular}} & \textbf{\begin{tabular}[c]{@{}l@{}}Semi-Sup.\\ \& Un-Sup.\end{tabular}} & \textbf{\begin{tabular}[c]{@{}c@{}}Statistical \\ Modeling\end{tabular}} & \textbf{\begin{tabular}[c]{@{}c@{}}Meta-Heuristic \\ Optimization\end{tabular}} & \textbf{\begin{tabular}[c]{@{}c@{}}Advanced Learning \\ Methods\end{tabular}} \\
\midrule
\textbf{\begin{tabular}[c]{@{}c@{}}Computational \\ Efficiency\end{tabular}}            & \begin{tabular}[c]{@{}l@{}}Training Time\end{tabular}                & \checkmark *                                                                                            & \checkmark                   & \checkmark                             & \checkmark                                   & \checkmark *                                \\            & \begin{tabular}[c]{@{}l@{}}Memory Usage\end{tabular}                 & \checkmark *                & \checkmark                   & \checkmark                             & \checkmark                                    & \checkmark *                                \\
                                                                                                         & \begin{tabular}[c]{@{}l@{}}Model Size\end{tabular}                   & \checkmark *                                                                                            & \checkmark                   & \checkmark                             & \checkmark                                    & \checkmark *                                \\
                                                                                                         & \begin{tabular}[c]{@{}l@{}}Model Complexity\end{tabular}             & \checkmark *                                                                                            & \checkmark *                 & \checkmark *                           & \checkmark *                                  & \checkmark *                            \\
                                                                                                         
\hline \textbf{\begin{tabular}[c]{@{}c@{}}Model Analysis \\\& Understanding\end{tabular}} & \begin{tabular}[c]{@{}l@{}}Feature Importance\end{tabular}           & \checkmark                                                                                              & -                   & \checkmark                             & -                                    & \checkmark                                  \\
                                                                                                         & \begin{tabular}[c]{@{}l@{}}Model Explainability\end{tabular}         & \checkmark                                                                                              & -                   & \checkmark                             & -                                    & \checkmark                                  \\
                                                                                                         & Interpretability                                                        & \checkmark                                                                                              & -                   & -                             & -                                    & \checkmark                                  \\
\hline \textbf{\begin{tabular}[c]{@{}c@{}}Data \\ Availability\end{tabular}}                  & \begin{tabular}[c]{@{}l@{}}Labeled Data\end{tabular}                 & \checkmark                                                                                              &  \checkmark                   & \checkmark                             & \checkmark                                    & \checkmark                                  \\
                                                                                                         & \begin{tabular}[c]{@{}l@{}}Unlabeled Data\end{tabular}               & -                                                                                              & \checkmark                   & -                             & -                                    & \checkmark                                  \\
\hline \textbf{\begin{tabular}[c]{@{}c@{}}Model \\ Performance\end{tabular}}                   & \begin{tabular}[c]{@{}l@{}}Performance on Large Data\end{tabular} & \checkmark                                                                                              & -                   & \checkmark                             & \checkmark *                                  & \checkmark                                  \\
                                                                                                         & \begin{tabular}[c]{@{}l@{}}Ability to Handle Noise\end{tabular}    & \checkmark                                                                                              & -                   & \checkmark                             & \checkmark *                                  & \checkmark                                  \\
                                                                                                         & Generalization                                                          & \checkmark                                                                                              & \checkmark                   & \checkmark                             & \checkmark                                    & \checkmark           \\
\bottomrule
\end{tabular} 

\footnotesize{$^*$includes variability- depends on specific factors such as the problem domain, dataset, or implementation}

\end{table}

Explanations of the considered algorithmic indicators are as follows, which we derived from the comprehensive analysis of the papers reviewed in this study.

\begin{enumerate}
\item Training time: Indicates the time required to train the model.
\item Memory usage: Reflects the memory required for training and storing the model.
\item Model size: Represents the model's size in terms of parameters.
\item Feature importance: Refers to the ability of the algorithm to provide insights into the importance of different features.
\item Model explainability: Indicates the ease of understanding the model's outcomes.
\item Generalization: Reflects the model's ability to perform well on unseen or test data.
\item Model complexity: Describes the complexity of the model in terms of its structure or mathematical formulation.
\item Labeled data: Indicates that the algorithms require labeled data for training, which is typical for supervised and semi-supervised learning tasks.
\item Unlabeled data: Leverage unlabeled data in unsupervised learning scenarios.
\item Performance on large data: Suggests that the algorithms perform well on large datasets.
\item Ability to handle noise: Can handle noise effectively, making them robust to noisy data.
\item Interpretability: Enabling a better understanding of the model's decision process.
\end{enumerate}


\begin{table}[!ht]
\scriptsize
\caption{Algorithmic approaches vs. rarity groups, modality, downstream tasks, and data processing tasks.$^*$}
\label{tab:algorithms2}
\begin{tabular}{lllllllll}

\toprule
\textbf{Algo. Group}                                                                                                     & \textbf{\begin{tabular}[c]{@{}l@{}}Sub Algo. \\ Group\end{tabular}}                    & \textbf{\begin{tabular}[c]{@{}l@{}}Algo. \\ Approach\end{tabular}}                                                                                                                                               & \textbf{Papers}                                                                                                                                                                                                                                            & \textbf{\begin{tabular}[c]{@{}l@{}}Rarity \\ Group\end{tabular}} & \textbf{\begin{tabular}[c]{@{}l@{}}Downstream \\ Tasks\end{tabular}} & \textbf{Modality} & 
\textbf{\begin{tabular}[c]{@{}l@{}}Dataset \\ Type\end{tabular}} & \textbf{\begin{tabular}[c]{@{}l@{}}Data Processing\\ Tasks\end{tabular}} \\
\midrule
\textbf{\begin{tabular}[c]{@{}l@{}}Supervised \\ Classification/\\ Regression \\ Methods\end{tabular}} & \begin{tabular}[c]{@{}l@{}}Threshold \\ methods\end{tabular}          & \begin{tabular}[c]{@{}l@{}}Logistic \\ regression\end{tabular}                                                                                                                                                   & 
\begin{tabular}[c]{@{}l@{}}
\cite{van2006prediction, zhao2018framework, olmucs2022comparison}, \\  \cite{alestra2014rare, maalouf2018logistic, fathy2020learning}  
\end{tabular}
& R4, R1                                                           & CF, FT                                                               & I,TX, N           & RE, DE, SIY            & SL, FE, DC, FS                                          \\&                                                                                        & \begin{tabular}[c]{@{}l@{}}Bayes \& \\ Neural networks\end{tabular}                                                                                                                                              & 
                    \begin{tabular}[c]{@{}l@{}}                                   \cite{bai2022rare, li2017rare, abbasi2022large, rafsunjani2019empirical},\\ \cite{bhanja2022black, yang2022applications,  parsa2021deep, lee2021early} \end{tabular}                                                                                           & R1,R2,R3,R4                                                      & CF                                                                   & N, A, I           & SI, RE, SIY  & SL, FE, DC, FS  \\
                                                                                                                         &                                                                                        & Autoencoders                                                                                                                                                                              &
\begin{tabular}[c]{@{}l@{}}\cite{bhanja2022black, martello2021improving, xu2022training},\\ \cite{ranjan2020understanding, hsieh2019unsupervised}
\end{tabular} & R1 & CF   & N   & RE  & SL, FE, DC, FS    \\
            \cline{2-9}
& \begin{tabular}[c]{@{}l@{}}Tree-based \\ classification \\ methods\end{tabular}        & Random Forest                               &
\begin{tabular}[c]{@{}l@{}}
\cite{coley2023empirical,bai2022rare, gondek2016prediction}, \\ \cite{fathy2020learning, hebert2016predicting,rafsunjani2019empirical}
\end{tabular}

& R1, R2                                                           & CF                                                                   & N                 & RE                    & FE, DC, FS                                                               \\
&                                                                                        & \begin{tabular}[c]{@{}l@{}}Boosted \\ Classification \\ Trees(XGBoost, \\ Adaboost)\end{tabular}                                                                                                                 & 
\begin{tabular}[c]{@{}l@{}}
\cite{hebert2016predicting, fathy2020learning, ashraf2023identification}, \\ \cite{rafsunjani2019empirical, ranjan2018dataset}   
\end{tabular}
& R1, R2, R3                                                       & CF                                                                   & N                 & RE                    & SL, FE, DC, FS                                                           \\
 
                                                     \\
\cline{2-9}                                                                                                                         & \begin{tabular}[c]{@{}l@{}}Cost-sensitive \\ learning\end{tabular}                     & \begin{tabular}[c]{@{}l@{}}Decision tree based \\ cost-sensitive learning, \\ Weighting based \\ cost-sensitive learning\end{tabular}                                                                            & \cite{zhao2018framework, li2017rare, he2021weighting, chen2004using}                                                                                                                                                                      & R1,R2,R3,R4                                                      & CF                                                                   & N, TX             & RE, SIY, DE            & SL, DC, FS                                                               \\
    \cline{2-9}                                                                                                                     & \begin{tabular}[c]{@{}l@{}}Non-parametric \\ classification \\ algorithms\end{tabular} & \begin{tabular}[c]{@{}l@{}}k-nearest neighbors \\ (k-NN)\end{tabular}                                                                                                                                            & \cite{rafsunjani2019empirical}                                                                                                                                                                                                            & R1, R2                                                           & CF                                                                   & N                 & RE                    & FE, DC                                                                   \\
    \cline{2-9}
            & \begin{tabular}[c]{@{}l@{}}Kernal-based \\ Methods\end{tabular}                        & SVM  & \begin{tabular}[c]{@{}l@{}}                                        \cite{iyer2015statistical, ahmadzadeh2019rare, alestra2014rare, zhao2018framework}, \\ 
            \cite{ling2021support, tang2008svms, maalouf2011robust, rafsunjani2019empirical}   \end{tabular}                                                                           & R1, R4, R2, R3                                                   & \begin{tabular}[c]{@{}l@{}}CF, \\ FT\end{tabular}                    & N, TX             & RE, SIY                & SL, DC, FS                                                               \\
\cline{2-9}                                                                                                                         & \begin{tabular}[c]{@{}l@{}}Inference/\\ Rule-Based \\ Methods\end{tabular}             & \begin{tabular}[c]{@{}l@{}}Inference methods,\\ More appropriate \\ inductive bias,\\ Two phase \\ rule induction,\\ Utilizing knowledge \\ and human \\ interactions,\\ Association \\ rule mining\end{tabular} & 
\begin{tabular}[c]{@{}l@{}}
\cite{cheon2009bayesian, neuman2021extreme, xiu2021variational}, \\ \cite{maalouf2011rare, weiss2004mining, kubat1998machine},\\ \cite{joshi2001mining, king2001explaining, yang2022applications},\\ 
\cite{parsa2021deep, cheon2009bayesian, berberidis2007detection}
\end{tabular}
                                                                & R1, R2, R3                                                       & \begin{tabular}[c]{@{}l@{}}CF, \\ FT\end{tabular}                    & N, I              & RE, DE                & SL, DC, FE                                                               \\
                                                                \cline{1-9}
\textbf{\begin{tabular}[c]{@{}l@{}}Semi-Supervised \\ \& Unsupervised\\ Methods\end{tabular}}                                                    &                          \multicolumn{2}{l}{\begin{tabular}[c]{@{}l@{}}Random forest, PAM, K-means,\\ Hierarchical, K-Nearest Neighbor,\\ BIRCH, K-Medoids \end{tabular}}                                              & 
\begin{tabular}[c]{@{}l@{}}
\cite{omar2022exploring, hamaguchi2019rare, wu2007local}, \\ \cite{rehab2021bearings, xu2022training, meng2006rare, nugraha2020clustering}
\end{tabular} 
& R1, R2, R3, R4                                                   & CL                                                                   & N, I              & RE, DE                & SL, FE, DC, FS                                                           \\ \cline{1-9}
\textbf{\begin{tabular}[c]{@{}l@{}}Statistical/ \\ Time series \\ Modeling\end{tabular}}                 & 
\multicolumn{2}{l}{\begin{tabular}[c]{@{}l@{}}Gumbel copula function \end{tabular}}  

 & \cite{yang2022prediction}  & R1                                                               & CF                                                                   & N                 & RE                    & FE, DC, FS                                             \\
&   
\multicolumn{2}{l}{\begin{tabular}[c]{@{}l@{}}ARIMA \end{tabular}} 
  & \cite{alestra2014rare} & R1 & \begin{tabular}[c]{@{}l@{}}CF, FT\end{tabular}                    & N                 & DE  & FE, DC, FS \\     &  
  \multicolumn{2}{l}{\begin{tabular}[c]{@{}l@{}}VAR \end{tabular}} 
  & \cite{hsieh2019unsupervised}  & R1, R2, R3, R4                                                   & CF                                                                   & N                 & RE  & FE, DC, FS                                                                                            \\
\cline{1-9}
\textbf{\begin{tabular}[c]{@{}l@{}}Meta-\\ Heuristic \\ Optimization\end{tabular}}                      & 
\multicolumn{2}{l}{\begin{tabular}[c]{@{}l@{}}
Particle swarm optimization, \\
Bat algorithm, \\
Genetic algorithms, \\
Evolutionary ensemble algorithms
\end{tabular}} & \cite{li2017adaptive, weiss1998learning, krawczyk2014cost}                                                                                                                                                                                                                     & R1, R2, R3, R4                                                       & CF                                                                   & N                 & RE , DE                   & SL                                                      \\
                                                                      
\cline{1-9}
\textbf{\begin{tabular}[c]{@{}l@{}}Advanced \\ Learning \\ Methods\end{tabular}}                        & 
\multicolumn{2}{l}{\begin{tabular}[c]{@{}l@{}}
Attention-based mechanisms
\end{tabular}}
& \cite{liu2021machinery, xu2021two, ravindranath2020m2nn, kulkarni2020soft}                                                                                                                                                                & R1, R2, R3, R4                                                   & CF                                                                   & N                 & RE                    & SL, FE, DC, FS                                          \\
                                                                          \cline{2-9}                                               & \begin{tabular}[c]{@{}l@{}}Markov \\ methods\end{tabular}                              & \begin{tabular}[c]{@{}l@{}}Extensible markov \\ models,\\ Monte carlo methods\end{tabular} & 
                                                                                                                         \begin{tabular}[c]{@{}l@{}}
                                                                                                                         \cite{meng2006rare, dang2022parallel, githubChaoDangRareEventEstimation},\\ \cite{olmucs2022comparison, bai2022rare, ling2021support, dangut2022application}  
                                                                               \end{tabular}                                          & R1, R2, R3, R4                                                   & CF, FT, RG                                                           & N                 &
                                                                               \begin{tabular}[c]{@{}l@{}}RE, SIY, \\ DE  
                                                                               \end{tabular}
                                                                               &       SL, FE, DC, FS                                                                   \\
\cline{2-9}        & 

\multicolumn{2}{l}{\begin{tabular}[c]{@{}l@{}}
Active and Meta learning\end{tabular}}                                                                                                          & 
        \begin{tabular}[c]{@{}l@{}}
\cite{dhulipala2022active, pickering2022discovering, lan2010study}, \\ \cite{gujrati2007meta, shi2020few}   
        \end{tabular}
        & R1                                                               & FT, SM, CF                                                           & N, I              & SIY, RE                &     SL, FE, DC, FS                                                                \\
        \bottomrule 
\end{tabular}

\footnotesize{$^*$N-Numeric, TX-Textual, I-Image, A-Audio, T-Time series, FT-Forecasting, CL-Clustering, CF-Classification, RG-Regression, RE-Naturally rare, DE-Derived, SIY-Simulated/Synthetic, SL-Sampling, FE-Feature engineering, DC-Data cleaning, FS-Feature selection}
\end{table}

This section extensively examined different algorithmic approaches applied in rare event prediction. 
The findings contribute to a better understanding of the diverse strategies employed in rare event prediction research and highlight the importance of selecting appropriate algorithms based on specific downstream tasks and modality requirements.

\clearpage

\section{Evaluation Approaches}
As ML models are developed, measuring the performance of every model is critical. Multiple evaluation metrics are employed in ML depending on the model and the results produced. This section focuses on the varied evaluation aspects used in rare event prediction. Firstly, we will highlight the significance of evaluation concerning rare event studies, followed by the evaluation methodologies and performance metrics used in related studies.

\begin{figure}[h]
  \centering
  \includegraphics[width=\linewidth]{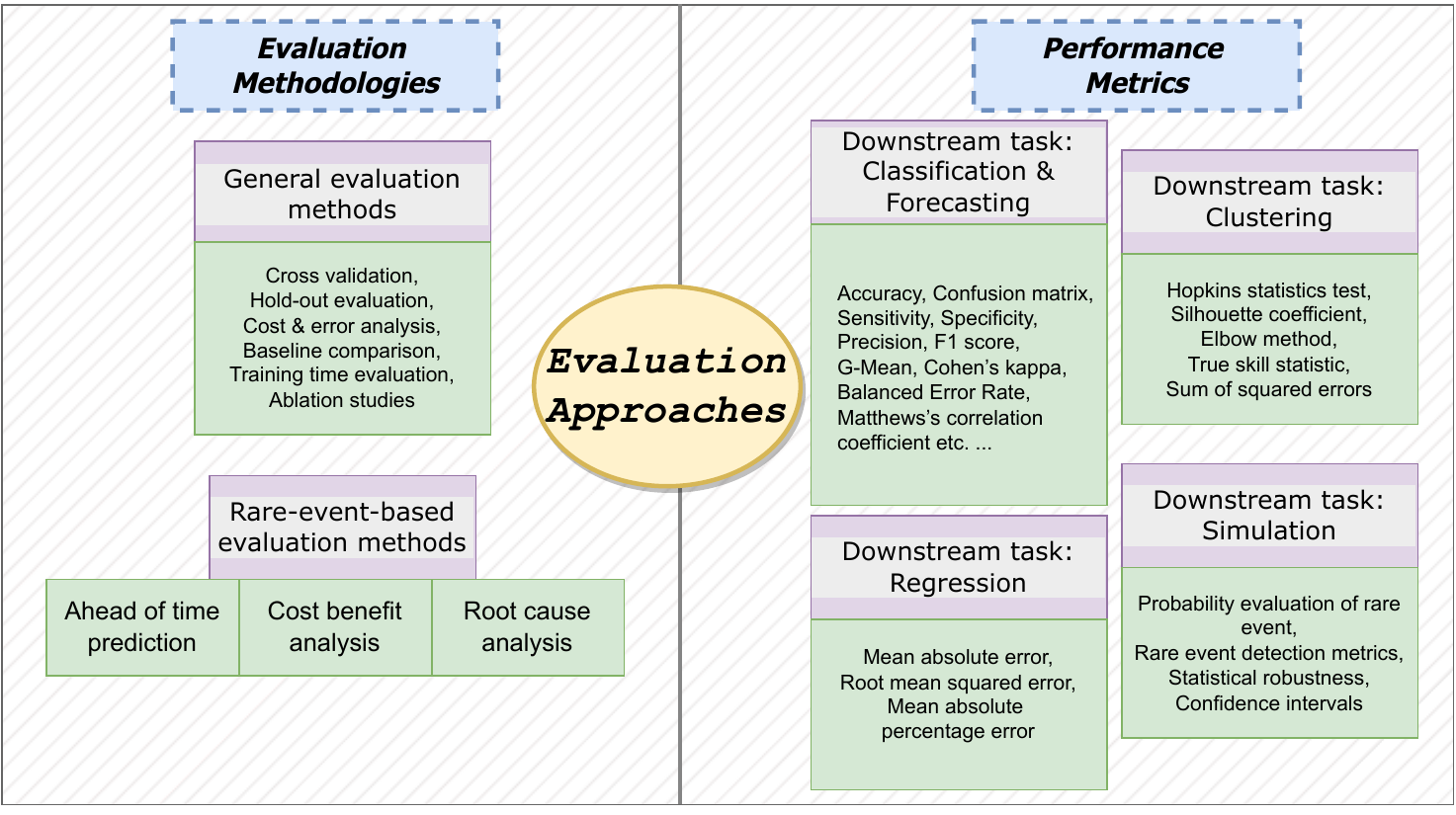}
  \caption{Evaluation approaches in rare-event prediction}
  \label{fig:evaluation}
\end{figure}

\subsection{Significance of evaluation}
Evaluation in rare event prediction research plays an integral role as it allows the assessment of the model's performance in accurately identifying and predicting rare events. Due to the imbalanced nature of rare events, traditional evaluation metrics might not provide a comprehensive assessment, making it vital to employ specialized evaluation techniques that focus on both overall performance and the ability to detect rare occurrences. Figure \ref{fig:evaluation} depicts the evaluation approaches we identified in rare event literature grouped into their evaluation categories.

\subsection{Evaluation Methodologies}
Evaluation methodologies in rare event prediction research involve general and rare event-specific approaches. General methods find widespread application in diverse ML tasks, while rare event-specific metrics are tailored to address the unique challenges and requirements of predicting rare events.

\subsubsection{General evaluation methods}
In rare event prediction research, various general evaluation methods are employed to assess the performance of algorithmic approaches (Table \ref{tab:evaluation1}). Cross-validation techniques, such as K-Fold, Stratified K-Fold, and Leave-One-Out are commonly used for performance estimation \cite{ahmadzadeh2019rare, Ranjan2019DataCD, chen2004using, coley2023empirical, fathy2020learning, gondek2016prediction, ali2014dynamic, kubat1998machine, zhao2018framework, hebert2016predicting, lee2021early}. Holdout evaluation through random splitting \cite{coley2023empirical, hebert2016predicting, liu2021machinery, ranjan2018dataset} and time-based splitting \cite{cheon2009bayesian, coley2023empirical, dangut2022application} is utilized to create distinct train-test datasets, facilitating the process of model training and evaluation. Analysis with standard baselines, cost and error analysis, training time evaluation \cite{bhanja2022black, bhanja2022black}, and ablation studies \cite{hamaguchi2019rare} for understanding individual components' contributions are among the general evaluation approaches employed in this field.

\begin{table}[!ht]
\caption{General evaluation methods in rare event evaluations.$^*$}
\label{tab:evaluation1}
\scriptsize

\begin{tabular}{lllllll}
\toprule
\textbf{Evaluation methodology} &
  \textbf{Sub methods} &
  \textbf{Papers} &
  \textbf{Rarity group} &
  \textbf{Algorithmic approach} &
  \textbf{Modality} &
  \textbf{\begin{tabular}[c]{@{}l@{}}Type of the \\ dataset\end{tabular}} \\
  \midrule
Cross validation methods &
  K-Fold Cross-Validation &
  \begin{tabular}[c]{@{}l@{}}\cite{ahmadzadeh2019rare, chen2004using, coley2023empirical, fathy2020learning}, \\
  \cite{gondek2016prediction, ali2014dynamic, kubat1998machine, wong2016statistical}\end{tabular} &
  R1, R2 &
  \begin{tabular}[c]{@{}l@{}}SVM, Cost-sensitive \\ learning, RF, LR,\\ XGBoost, \\ IBL \end{tabular} &
  N, I, TX &
  DE, RE \\ \\
 &
  LOOCV , Stratified K-Fold &
  \begin{tabular}[c]{@{}l@{}}\cite{kubat1998machine, zhao2018framework, hebert2016predicting, lee2021early}\end{tabular} &
  R1, R2, R4 &
  \begin{tabular}[c]{@{}l@{}}Cost-sensitive learning, \\ IBL, LR, SVM, RF,\\ XGBoost, CNN, \\ VAR, k-NN\end{tabular} &
  I, N, TX &
  RE \\ 
\\
\begin{tabular}[c]{@{}l@{}}Holdout evaluation\\ (Train-test-validation splitting)\end{tabular} &
  Random Split &
  \begin{tabular}[c]{@{}l@{}}\cite{coley2023empirical,  hebert2016predicting, liu2021machinery, ranjan2018dataset}\end{tabular} &
  R1, R4 &
  \begin{tabular}[c]{@{}l@{}}
  RF, XGBoost, \\ Attention-based, \\ Adaboost 
  \end{tabular} &
  N &
  RE \\
 &
  Time-Based Split &
  \begin{tabular}[c]{@{}l@{}}\cite{cheon2009bayesian,coley2023empirical,dangut2022application}\end{tabular} &
  R1,R3,R4 &
  \begin{tabular}[c]{@{}l@{}}Bayesian methods, RF,\\ Monte Carlo methods\end{tabular} &
  N &
  DE, RE \\
  \begin{tabular}[c]{@{}l@{}}
Cost and error analysis, \\
Baseline comparison \end{tabular} &
   &
   \cite{abbasi2022large, bhanja2022black, cheon2009bayesian, dangut2022application}
   & R1, R2, R3, R4
   & \begin{tabular}[c]{@{}l@{}} SVM, KNN, XGB,\\ MLP, RF, LR \end{tabular}
   & \begin{tabular}[c]{@{}l@{}} N, A, \\ TX, I \end{tabular}
   & \begin{tabular}[c]{@{}l@{}} RE, DE, \\ SI, SY \end{tabular}
   \\

Training time evaluation & &
  \cite{bhanja2022black, lee2021early} &
  R1 &
  \begin{tabular}[c]{@{}l@{}}
  CNN-Autoencoders,\\ CNN, VAR, \\ k-nearest neighbors
  \end{tabular}
  &
  N &
  RE \\
Ablation studies &
   &
  \cite{hamaguchi2019rare} &
  R1, R4 &
  K-means, One-class learning &
  I, N &
  DE\\
  \bottomrule

\end{tabular}

\footnotesize{$^*$N-Numeric, TX-Textual, I-Image, A-Audio, T-Time series, RE-Naturally rare, DE-Derived, SI-Simulated, SY-Synthetic}
\end{table}

\subsubsection{Rare event-specific evaluation methods}
Evaluation methods that are specific to rare events emphasize early prediction, proactive decision-making, and economic feasibility in the prediction task. We identified several methods below as specific to rare events as presented in Table \ref{tab:evaluation2}.

\noindent \textbf{\textit{I) Ahead-of-time prediction evaluation}}:
Ahead-of-time prediction evaluation assesses a model's predictive capability for events or outcomes occurring multiple time steps into the future. This approach contrasts with the usual prediction evaluation, which focuses on predicting the immediate next event or outcome. Evaluating models "ahead of time" is particularly important in scenarios where early detection and proactive decision-making are critical. In the studies by \cite{bhanja2022black, ranjan2020understanding, dangut2022application}, the performance of the proposed models has been evaluated over various lead times, ranging from one to many units of time. It was observed that while the accuracy of the models might decrease as the lead time increases, their performance has remained stable, demonstrating their potential utility in providing valuable insights well before the occurrence of rare events, such as black-swan or equipment failures.

\noindent \textbf{\textit{II) Cost-benefit analysis}}:
Cost-benefit analysis has been used to assess rare event prediction's economic feasibility and potential advantages. Methods like assessing potential financial loss reduction, optimizing resource allocation, and evaluating the potential mitigation of losses or damages are utilized to quantify and compare the costs and benefits linked to early prediction of rare events, as explored in \cite{lee2021early} and \cite{neuman2021extreme}.

\noindent \textbf{\textit{III) Root cause analysis}}:
This could be viewed as a problem-solving approach focused on discerning the fundamental elements contributing to rare occurrences. It involves investigating and tracing the chain of events to determine the fundamental causes \cite{lee2021early}.

\begin{table}[!ht]
\caption{Rare event-based evaluation methods.$^*$}
\label{tab:evaluation2}
\resizebox{\columnwidth}{!}{%
\begin{tabular}{llllll}
\toprule
\textbf{Evaluation methodology} &
  \textbf{Papers} &
  \textbf{Rarity group} &
  \textbf{Algorithmic approach} &
  \textbf{Modality} &
  \textbf{\begin{tabular}[c]{@{}l@{}}Type of the \\ dataset\end{tabular}} \\ 
  \midrule
\begin{tabular}[c]{@{}l@{}}Ahead-of-time prediction \\ evaluation\end{tabular} &
  \begin{tabular}[c]{@{}l@{}}\cite{bhanja2022black, dangut2022application, weiss1998learning, xu2022training}\end{tabular} &
  R1, R3 &
  \begin{tabular}[c]{@{}l@{}}CNN-Autoencoders, CNN, \\ Genetic algorithms\end{tabular} &
  N &
  RE, DE \\
Cost-benefit analysis &
  \cite{lee2021early, neuman2021extreme} &
  R1 &
  \begin{tabular}[c]{@{}l@{}}k-nearest neighbors,\\ Bayes and neural networks,\\ Bayesian methods\end{tabular} &
  N &
  RE \\
Root cause analysis &
  \cite{lee2021early} &
  R2 &
  \begin{tabular}[c]{@{}l@{}}k-nearest neighbors,\\ Bayes and neural networks\end{tabular} &
  N &
  RE \\
  \bottomrule
\end{tabular}
}
\footnotesize{$^*$N-Numeric, TX-Textual, I-Image, A-Audio, T-Time series, RE-Naturally rare, DE-Derived, SI-Simulated, SY-Synthetic}
\end{table}

\subsection{Performance metrics}
Performance metrics aid in monitoring and optimizing algorithmic approaches, facilitating comparison of algorithms, determining optimal thresholds, supporting data-driven decision-making and conducting risk assessments. We categorize these performance metrics used in rare event research, which differ based on the downstream tasks involved. The detailed analysis of performance metrics and rarity groups, algorithmic approaches, modality, and dataset types is included in Table \ref{tab:performancemetrics}.

\textbf{\textit{I) Downstream tasks: Classification and Forecasting}}
\\
\textbf{\textit{i) Accuracy}}: Evaluating model accuracy is essential, particularly in imbalanced scenarios, where relying solely on accuracy can be deceptive. Since accuracy only assigns overall class weights rather than weights for unusual classes or minorities. Zhao et al. \cite{zhao2018framework} highlight that standard classifiers like LR and SVM assume equal class distribution, which leads to poor sensitivity for rare events. To address this, variants of accuracy such as Faulty-Normal Accuracy (FNACC) and Real Faulty-Normal Accuracy (RFNACC) are introduced \cite{alestra2014rare, marins2021fault}. FNACC assesses how well the model identifies normal samples preceding faults, while RFNACC evaluates the accuracy in predicting actual faults, offering a more nuanced view of model performance on rare events.


\textbf{\textit{ii) Geometric mean (G-Mean)}}: G-Mean combines sensitivity and specificity to measure the balance between classification performance for the majority and minority classes. It is particularly useful in rare event prediction as a low G-Mean indicates poor prediction of rare events, despite good performance on the majority class \cite{chen2004using, wang2012applying, tang2008svms}.

\textbf{\textit{iii) Cohen’s kappa index}}: It is a metric that has been used in earth science and healthcare-based rare event prediction evaluation for use cases like landslide prediction\cite{van2006prediction},  hazardous seismic bumps in coal mines, detecting changes in geospatial trajectories \cite{iyer2015statistical}, \cite{li2017rare}, mineral prospectivity prediction \cite{yang2022applications} and thoracic events prediction \cite{li2017adaptive}.  \cite{li2017rare} has selected Kappa as the primary evaluation metric since it indicates the generalizability of the classifier's predictive ability on the supplementary datasets. However, some research attempts have proven that the Kappa metric produces unreliable results due to its high sensitivity to the distribution of the marginal totals \cite{chicco2020advantages}.

\textbf{\textit{iv) Balanced error rate (BER)}}: BER measures the average of the errors for each class in a classification problem, which has been used by \cite{li2017rare, iyer2015statistical} in rare event model evaluation. 

\textbf{\textit{v) Matthews’s correlation coefficient (MCC)}}: MCC is a more accurate statistical metric that provides a reliable measure of a model's performance by considering all four confusion matrix categories (TP, FN, TN, and FP) and balancing the impact of both positive and negative class sizes \cite{chicco2020advantages, li2017rare, yang2023extreme}.
Hence, this metric is beneficial when dealing with highly imbalanced rare event datasets.


\textbf{\textit{vi) Cost matrix}}: A cost matrix specifies the relative importance of accuracy for different classification predictions by assigning costs or weights to various classification predictions. In a cost matrix, positive numbers (costs) influence negative outcomes, whereas negative numbers (benefits) influence positive outcomes. An instance is \cite{fathy2020learning}, which tries to minimize the cost matrix for binary misclassification using threshold and tree-based classification models. This approach helps prioritize detecting rare events by adjusting the cost of false positives and false negatives.

\textbf{\textit{vii) Top-decile lift (TDL)}}:
TDL is a metric used in the economic domain that compares the incidence in the top 10\% samples with the highest model predictions to the incidence of the entire sample \cite{berry2004data}.  \cite{ali2014dynamic} used TDL and assumed it returned the customers predicted to be most likely to churn rather than a random selection. In rare event scenarios, TDL helps the model identify the most promising cases by highlighting its ability to distinguish rare events from non-rare ones, thus providing insights beyond overall accuracy.

\textbf{\textit{viii) Reconstruction error}}:
This measures the discrepancy between the original input data and its reconstructed version generated by a model by quantifying how well the model can reproduce the input data. In models like autoencoders, the model is trained to reconstruct its input data from a compressed latent representation \cite{hsieh2019unsupervised}.
Reconstruction error identifies significant deviations from normal patterns, aiding in effective rare event prediction and enhancing the model's ability to prioritize and identify rare events.

\textbf{\textit{ix) Moran index}}:
Used in spatial analysis and statistics and measures a variable's spatial autocorrelation or clustering within a geographic space. It calculates the degree of similarity or dissimilarity between neighboring locations based on the values of the analyzed variable. It provides a measure of spatial dependence, indicating whether similar values tend to cluster together (positive spatial autocorrelation) or are dispersed (negative spatial autocorrelation) \cite{yang2022prediction}. Moran index has been used to detect spatial clustering of rare events, and it gives insights into spatial locations where these events are more likely to occur \cite{rumi2018crime,yang2022prediction}.

\noindent \textbf{\textit{II) Downstream task: Clustering}}

\textbf{\textit{i) Elbow method}}:
The Elbow method, typically used to identify the optimal number of clusters, is beneficial in rare event prediction tasks where the majority class can dominate the clustering process. This also ensures that minority class clusters are not merged into larger majority class clusters, thereby preserving the distinct characteristics of rare events. This approach is helpful in studies focusing on class imbalance, where detecting minority classes with high specificity and sensitivity is vital for model performance \cite{nugraha2020clustering}.

\textbf{\textit{ii) Silhouette coefficient}}:
It is another metric used for internal cluster validation. This metric is particularly important in rare event prediction as it helps to determine whether rare event clusters are well-defined and distinct from non-rare event clusters. A small silhouette width may indicate that rare events are being inaccurately grouped with non-rare events, which can significantly impact the model's ability to identify and predict rare occurrences \cite{omar2022exploring}.

\textbf{\textit{iii) Hopkins statistics test}}
The Hopkins statistics test is used to assess the clustering tendency of a dataset. Omar et al. \cite{omar2022exploring} used it to evaluate the clustering tendency in a rare event dataset. This revealed well-defined and meaningful clusters that standard clustering evaluation methods might miss due to the high data imbalance and the distinct patterns often exhibited by the minority class.

\textbf{\textit{iv) True skill statistic (TSS)}}:
TSS combines sensitivity and specificity into a single measure, offering a more balanced model performance evaluation in rare event prediction. Unlike accuracy, which can be misleading in imbalanced datasets, TSS provides a robust assessment of a model's positive and negative predictive abilities \cite{yoon2023application, allouche2006assessing}. This makes it particularly useful in scenarios like species distribution modeling or solar flare forecasting, where both false positives and false negatives carry significant consequences \cite{ahmadzadeh2019rare}.

\textbf{\textit{v) Sum of squared errors (SSE)}}:
SSE measures the variation within clusters by calculating the sum of the squared differences between each observation and the group mean and is often used to evaluate cluster compactness \cite{thinsungnoena2015clustering}. In rare event prediction, minimizing SSE ensures that rare event clusters are as distinct and well-defined as possible \cite{rehab2021bearings}. High SSE values may indicate that the model fails to accurately capture the unique characteristics of rare events, leading to poor prediction performance \cite{rehab2021bearings}.

\noindent \textbf{\textit{III) Downstream task: Regression}}

\textbf{\textit{i) Mean absolute error (MAE) and root mean squared error (RMSE),
mean absolute percentage error (MAPE)}}:

MAE or Mean Absolute Deviation (MAD), RMSE, and MAPE or Mean Absolute Percentage Deviation (MAPD) are popular scale-dependent metrics used in evaluation. MAE provides a straightforward measure of prediction error, highlighting performance on underrepresented rare events. RMSE penalizes more significant errors, making it ideal for assessing precision in datasets where rare occurrences have high variance. Though sensitive to small actual values, MAPE offers insights into percentage accuracy, which is essential for comparing model performance across different events. These metrics have been employed in rare event studies, with Xu et al. \cite{xu2021two} selecting MAE and RMSE as regression loss functions and Ravindranath et al. \cite{ravindranath2020m2nn} using these metrics in multivariate multiscale attention research.

\noindent \textbf{\textit{IV) Downstream task: Simulation}}

\textbf{\textit{i) Probability evaluation of rare event}}:
This metric directly measures the likelihood of the rare event occurring. It provides a quantitative assessment in a simulation of the occurrence of the event of interest \cite{bai2022rare, dhulipala2022active, olmucs2022comparison}.

\textbf{\textit{ii) Rare event detection metrics}}:
The standard metrics that evaluate the performance of algorithms designed to predict rare events like precision, recall, F1 score, and Area Under the Receiver Operating Characteristic Curve (AUC-ROC), RMSE, have also been utilized in simulation evaluations \cite{bai2022rare, olmucs2022comparison}.

\textbf{\textit{iii) Statistical robustness}}:
Statistical robustness refers to the ability of statistical models to produce reliable and consistent results under varying sample sizes, distributional assumptions, or data perturbations \cite{bai2022rare}.

\textbf{\textit{iv) Confidence intervals}}:
This metric provides a range of values within which the actual value of a performance metric is expected to fall and quantifies the uncertainty associated with the estimated performance metric \cite{olmucs2022comparison}.

\begin{table}[t]
\caption{Performance metrics used in Rare event evaluations.$^*$}
\label{tab:performancemetrics}
\scriptsize
\begin{longtable}{lllllll}
\toprule
\textbf{Downstream task} &
  \textbf{Performance metric} &
  \textbf{Papers} &
  \textbf{Rarity group} &
  \textbf{Algorithmic approach} &
  \textbf{Modality} &
  \textbf{\begin{tabular}[c]{@{}l@{}}Dataset type \end{tabular}} \\
\midrule
\endhead
\begin{tabular}[c]{@{}l@{}} Downstream task: \\ Classification and \\ Forecasting \end{tabular} &
  Accuracy &
  \begin{tabular}[c]{@{}l@{}}\cite{abbasi2022large, bhanja2022black,coley2023empirical}, \\ \cite{dangut2022application, rafsunjani2019empirical, wong2016statistical}\end{tabular} &
  R4, R1,R2 &
  \begin{tabular}[c]{@{}l@{}}Bayes \& Neural networks, RF,\\ Monte Carlo methods\end{tabular} &
  A, N, TX &
  RE, DE \\
 &
  \begin{tabular}[c]{@{}l@{}}Confusion matrix, Sensitivity, \\ Specificity, FPR, FNR, \\ Precision, F1 score\end{tabular} &
  \begin{tabular}[c]{@{}l@{}}\cite{abbasi2022large, berberidis2007detection, bhanja2022black}, \\ \cite{chen2004using, cheon2009bayesian, coley2023empirical}, \\ \cite{dangut2022application, rafsunjani2019empirical, hsieh2019unsupervised}, \\ \cite{kulkarni2020soft, weiss1998learning}, \\ \cite{wong2016statistical, zhao2018framework} \end{tabular} &
  R1, R3,R4, R2 &
  \begin{tabular}[c]{@{}l@{}}Bayes \& Neural networks,\\ Association rule mining,\\ Cost-sensitive learning,\\ Bayesian methods, RF,\\ Monte Carlo methods,\\ VAR, LSTM-autoencoder,\\ Genetic algos\end{tabular} &
  A, N, TX &
  RE, DE \\
 &
  Geometric Mean (G-Mean) &
  \begin{tabular}[c]{@{}l@{}}\cite{berberidis2007detection,  dangut2022application, kubat1998machine}\end{tabular} &
  R1, R3,R4, R2 &
  \begin{tabular}[c]{@{}l@{}}Association \\ rule mining,\\ Monte Carlo methods,\\ Attention-based mechanisms\end{tabular} &
  N &
  DE, RE \\
 &
  Cohen’s Kappa Index &
  \begin{tabular}[c]{@{}l@{}}\cite{iyer2015statistical, li2017adaptive, yang2022applications}\end{tabular} &
  R1, R2, R3 &
  \begin{tabular}[c]{@{}l@{}}SVM, Particle swarm optimization,\\ Bat algorithm, CNN\end{tabular} &
  N, I &
  Real \\
 &
  Balanced Error Rate &
 \cite{iyer2015statistical} &
   &
  SVM &
  N &
   \\
 &
  Matthews’s correlation coefficient &
 \cite{li2017rare} &
  R3 &
  \begin{tabular}[c]{@{}l@{}}Cost-sensitive learning\\ Bayes \& Neural networks\end{tabular} &
  N &
  RE \\
 &
  PR curve, ROC, AUC and AUPRC &
  \begin{tabular}[c]{@{}l@{}}\cite{abbasi2022large,  alestra2014rare, chen2004using}, \\ \cite{coley2023empirical, ali2014dynamic, wong2016statistical}\end{tabular} &
  R1, R4, R2 &
  \begin{tabular}[c]{@{}l@{}}Bayes \& Neural networks,\\ ARIMA, SVM, Cost-sensitive \\ learning, RF \end{tabular} &
  A, N, TX &
  DE, RE \\
 &
  Cost matrix &
 \cite{fathy2020learning} &
  R2 &
  RF, LR, XGB &
  N &
  RE \\
 &
  Top-decile lift &
 \cite{ali2014dynamic} &
   &
  Hybrid &
  N &
  RE \\
 &
  Reconstruction error &
 \cite{hsieh2019unsupervised, xu2022training} &
  R1 &
  \begin{tabular}[c]{@{}l@{}}VAR, LSTM-autoencoder,\\ autoencoders, hierarchical clustering\end{tabular} &
  N &
  RE \\
 &
  Moran index &
 \cite{yang2022prediction} &
  R1 &
  Statistical Modeling &
  N &
  RE \\
  \cline{1-7} 
\multicolumn{7}{l}{} \\
\begin{tabular}[c]{@{}l@{}}Downstream task: \\ Clustering\end{tabular} &
  Hopkins statistics test &
 \cite{omar2022exploring} &
  R2 &
  \begin{tabular}[c]{@{}l@{}}Random forest clustering,\\ PAM\end{tabular} &
  N &
  RE \\
 &
  Silhouette coefficient &
 \cite{omar2022exploring} &
  R3 &
  \begin{tabular}[c]{@{}l@{}}Random forest clustering,\\ PAM\end{tabular} &
  N &
  RE \\
 &
  Elbow method &
 \cite{nugraha2020clustering} &
  R1, R2 &
  K-Medoids clustering &
  N &
  RE \\
 &
  True Skill Statistic &
 \cite{ahmadzadeh2019rare} &
  R1, R2 &
  Kernel-based methods - SVM &
  N &
  DE \\
 &
  Sum of Squared Errors  &
 \cite{rehab2021bearings} &
  R4 &
  K-nearest neighbur &
  N &
  RE \\
  \cline{1-7} 
\multicolumn{7}{l}{} \\
\begin{tabular}[c]{@{}l@{}}Downstream task: \\ Regression\end{tabular} &
  \begin{tabular}[c]{@{}l@{}}Mean Absolute Error and \\ Root Mean Squared Error,\\ Mean absolute percentage error \end{tabular} &
 \cite{xu2021two, ravindranath2020m2nn} &
  R1, R3, R4 &
  \begin{tabular}[c]{@{}l@{}}Attention-based \\ mechanisms\end{tabular} &
  N &
  RE \\
  \cline{1-7} 
\multicolumn{7}{l}{} \\
\begin{tabular}[c]{@{}l@{}}Downstream task: \\ Simulation\end{tabular} &
  Probability evaluation of rare events &
  \begin{tabular}[c]{@{}l@{}}\cite{bai2022rare, dhulipala2022active, olmucs2022comparison}\end{tabular} &
  R1,R2,R3,R4 &
  \begin{tabular}[c]{@{}l@{}}Monte carlo methods,\\ Bayes and NN, RF, LR\end{tabular} &
  N &
  SI \\
 &
  Rare event detection metrics &
 \cite{bai2022rare, olmucs2022comparison} &
  R1,R2,R3,R4 &
  \begin{tabular}[c]{@{}l@{}}Monte carlo methods,\\ Bayes and NN, RF, LR\end{tabular} &
  N &
  SI \\
 &
  Statistical Robustness &
 \cite{bai2022rare} &
  R1,R2,R3,R4 &
  \begin{tabular}[c]{@{}l@{}}Monte carlo methods,\\ Bayes and NN, RF\end{tabular} &
  N &
  SI \\
 &
  Confidence intervals &
 \cite{olmucs2022comparison} &
  R1,R2,R3,R4 &
  LR, Monte carlo &
  N &
  SI \\
  \bottomrule
\end{longtable}
\footnotesize{$^*$N-Numeric, TX-Textual, I-Image, A-Audio, T-Time series, RE-Naturally rare, DE-Derived, SI-Simulated, SY-Synthetic}
\end{table}

\subsubsection{Analyzing evaluation approaches vs. rarity groups, modality, type of data and algorithmic approaches}

Based on our literature review, the analysis of general evaluation methods and rare event-specific evaluation methods, along with their respective components, is summarized in Tables \ref{tab:evaluation1} and \ref{tab:evaluation2}. The analysis highlighted that standard techniques like cross-validation have predominantly been employed in datasets encompassing numerical, textual, and image data, particularly in extremely-rare and very-rare categories. Only limited research has been dedicated to rare event-specific evaluation methods applied to numerical data. In classification and forecasting, widely used performance metrics such as accuracy, precision, recall, AUC, and ROC have been employed for evaluation, regardless of the data modalities. However, evaluation has predominantly been conducted on numerical data for other downstream tasks like clustering, regression, and simulation. Moreover, our analysis reveals the limited availability of rare event-specific evaluation techniques and highlights the need for conceptualized evaluation methodologies explicitly designed for rare event prediction. Such tailored approaches would be essential to address the unique challenges of rare events.

In conclusion, we presented a comprehensive analysis of various evaluation approaches used in rare event prediction.
From our analysis, we identified most studies rely on standard evaluation metrics, highlighting the need for tailored metrics explicitly designed for rare events. This is particularly important given the limited availability of rare event-specific evaluation techniques, especially in datasets comprising numerical, textual, and image data across diverse tasks such as classification, forecasting, clustering, regression, and simulation. Through this examination, we also gained valuable insights into the suitability and limitations of different evaluation methodologies and performance metrics for handling rare events. The findings underscore the significance of choosing appropriate metrics tailored to specific downstream tasks and modalities.

\section{Research Findings and Discussion}
This section provides an overview of the research findings, emphasizing identified gaps, and open challenges in the field. It also explores emerging research trends and novel approaches, aiming to address the complexities of rare event prediction and drive advancements in the field.

\subsection{The gaps in current literature}

\begin{enumerate}
\item Lack of standardized benchmark datasets: “Standardized dataset” represents real-world scenarios that include an ideal rarity percentage, annotated ground truth, scalability, temporal and spatial association, feature diversity, noise, and outliers. They should also adhere to privacy and ethical standards to enable fair and comprehensive comparisons of algorithms and advance the state-of-the-art in the field. The absence of such widely accepted benchmark datasets for rare event prediction makes it challenging to compare the performance of different algorithms and approaches consistently.
\item Imbalanced data processing techniques: While many studies explore standard methods for addressing imbalanced data, there is a necessity for in-depth investigation of standardized techniques for data processing, especially for handling extreme rarity, since general data imbalance techniques may inadequate and underperform.
\item Scalability issues: Current research has not addressed scalability challenges in handling large-scale rare event datasets, such as computational complexity, memory constraints, and algorithmic efficiency.
\item Limited focus on Uncertainty Quantification (UQ): UQ involves quantifying and characterizing uncertainties associated with predictions, offering insights into the reliability and confidence of estimates. The lack of comprehensive studies on UQ for rare event prediction undermines confidence in model predictions, which is essential for making informed decisions when dealing with infrequent and high-impact events.

\item Real-world applicability: More research is needed to assess the practical usability and robustness of prediction models in critical real-world applications, particularly in domains such as healthcare, finance, and earth sciences.
\item Limited focus on rare event-specific evaluation techniques: Many studies still rely on standard and general evaluation metrics designed for balanced datasets, which is challenging because these metrics fail to accurately reflect model performance in rare events, leading to misleading assessments and suboptimal model improvements.

\item Measuring data quality: The absence of standardized methods for measuring data quality in data sampling and augmentation for rare event data is a research gap.
\end{enumerate}

\subsection{Open challenges of rare event prediction}Based on our study, below are some open issues and challenges in the field of rare event prediction that we identified.
\begin{enumerate}
\item 
Lack of annotated data: Collecting and annotating data about rare events proves complex and time-intensive, posing an inherent obstacle in developing proficient rare event prediction models.

\item 
Interdependence between rare events: Identifying complex relationships and dependencies among rare events (when more than one rare event occurs) remains elusive.

\item 
Bias in data: Bias refers to systematic errors or prejudices in data collection, sampling, or labeling that can skew the results of prediction models. This bias is especially challenging in rare events since it can lead to significant misrepresentations of the events, resulting in highly inaccurate and unreliable predictions.

\item 
Generalization and real-world applicability: Achieving generalizability in rare event prediction models presents challenges due to the potentially significant differences
in variability and distribution of new, unseen, or real-world data compared with the training data.

\item 
Interpretability: The infrequency and high stakes of rare events necessitate clear and understandable explanations to ensure trust and accurate decision-making in critical domains like healthcare or finance.

\end{enumerate}

\subsection{Research trends in rare event prediction}

Research in rare event prediction has surged in recent years due to the increasing importance of rare events in many fields. Some of the research trends in this area include:

\begin{enumerate}

\item
Incorporation of domain knowledge: Domain knowledge, expert opinions, and human insights Human-in-the-Loop Systems) can aid in rare event prediction. They have led in emerging applications that combine data-driven and algorithmic approaches with qualitative knowledge, knowledge graphs, expert systems, and rule-based models to improve the robustness and reliability of predictions \cite{sanjak2024clustering, mitropoulou2024anomaly, xian2024physics, varsou2024deepprobcep, gama2024neuro}.

\item 
Explainability: The growing emphasis on transparency and comprehension in model predictions has led to the rise of explainability as a prospective future trend in rare event prediction research \cite{reimann2024predicting, reimann2024predicting, felix2024explainable}. This development aims to enhance decision-making processes and foster trust in the accuracy of models.

\item
Ensemble learning: This involves combining multiple algorithms to improve prediction performance, which is particularly effective in handling imbalanced datasets and has recently gained attention \cite{marel2024predicting, frazee2024deepdrivewe, dhaubhadel2024high, moura2024predicting}.

\item 
Meta-learning, few-shot learning, and transfer learning:
Meta-learning \cite{peng4684217forecasting} and few-shot learning \cite{zhang2024few} have proven to enhance rare event prediction recently by enabling model learning from related events and generalization to new and infrequent events with limited labeled data, thus improving adaptability and accuracy.
Transfer learning is a meta-learning technique that involves using knowledge from one domain to improve learning in another. This approach is promising in rare event prediction, where data from different domains can be leveraged to improve predictive performance \cite{dhaubhadel2024high, jacques2024estimation}.

\item
Using multi-modal data: Integrating information from multiple data sources and modalities has improved rare event predictive power, generalization, and robustness while uncovering hidden patterns \cite{arno2024numbers, ravindranath2024mma}. However, it also presents challenges related to data integration, feature engineering, and model complexity.

\item
Uncertainty quantification with rarity: Recently, UQ has gained prominence within rare event prediction. Using UQ, decision-makers can make more informed choices, assess risk, optimize resource allocation, and enhance the robustness to mitigate the impacts of rare events. These can incorporate advancements using probabilistic modeling \cite{xian2024physics}, Bayesian inference \cite{gong2024multifidelity}, Monte Carlo methods \cite{grewal2024predicting}, and Deep Ensembles \cite{kar2024xwavenet, grewal2024predicting}.
\item{Privacy-preserving techniques: Rare event research on techniques like federated learning, differential privacy, encrypted data processing \cite{jafarigol2024distributed, bukhari2024secure}, and privacy-preserving data mining \cite{gui2024privacy} can be prioritized for sensitive rare event data to ensure confidentiality }.
\item{Leveraging edge devices and neuromorphic computing techniques: Incorporating edge devices and advanced neuromorphic computing techniques has emerged as a trend in rare event prediction \cite{shaheen2024ai, ramya2024analysis}, enabling efficient real-time processing, low-latency analytics, and enhanced model adaptability at the edge while leveraging efficient computing paradigms for improved pattern recognition and prediction accuracy.
}
\end{enumerate}

\section{Vision Forward}
\begin{figure}[h]
  \centering
  \includegraphics[width=0.75\linewidth]{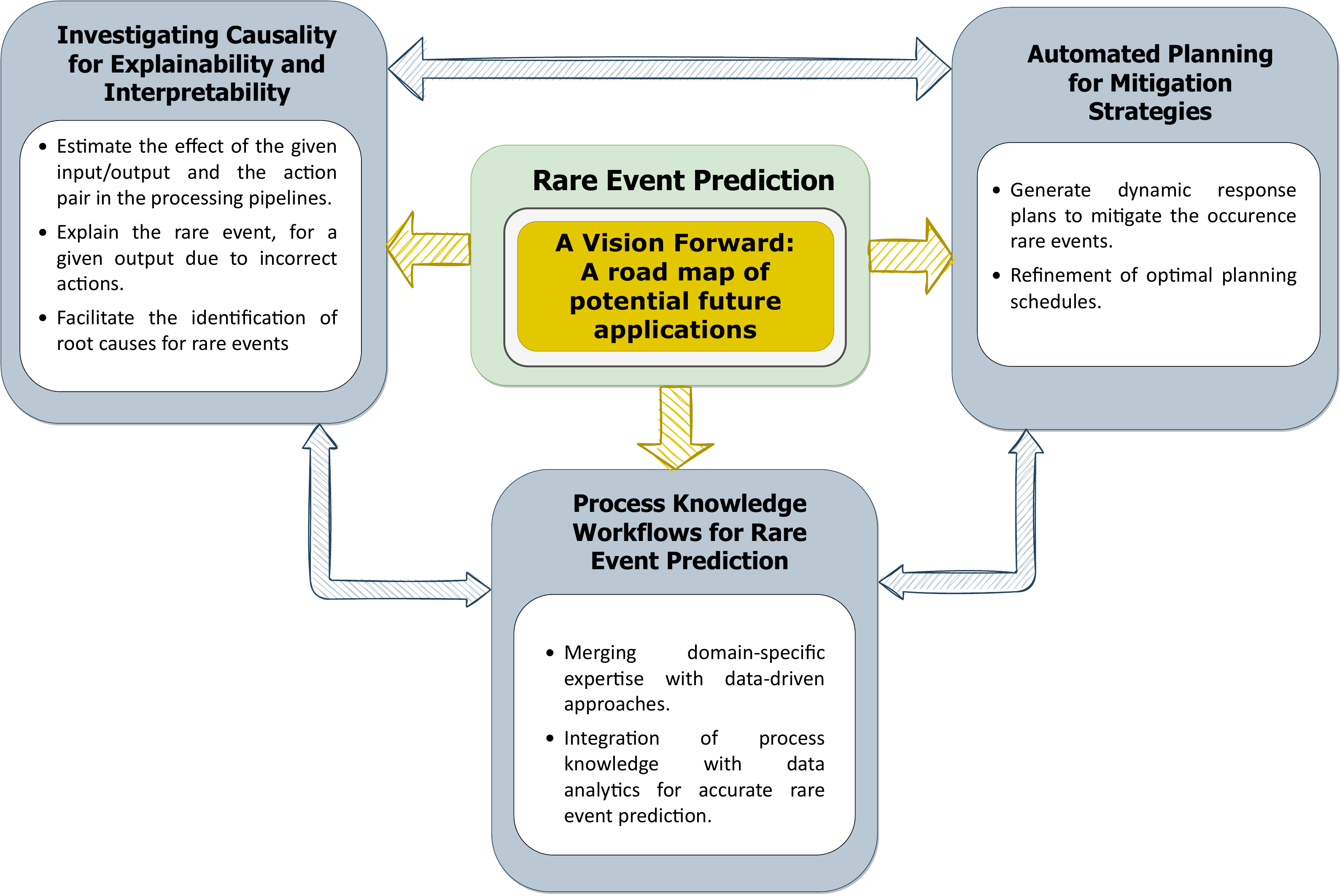}
  \caption{The road map of potential future applications in rare event prediction}
  \label{fig:roadmap}
\end{figure}

As the methods, data, and applications for predicting rare events continue to advance, we can see how it will affect tasks beyond prediction. We envision that it will have a significant impact on three specific areas: improving our understanding of causality, using process knowledge workflows to predict rare events, and implementing automated planning strategies to mitigate them effectively. This is illustrated in Figure \ref{fig:roadmap}.

\vspace{4mm}
\begin{enumerate}
    \item Investigating causality for explainability and interpretability

An integral component of our strategy entails delving deeper into causality, which will aid us in delivering more lucid explanations. This requires us to gauge the influence of individual input-output associations and actions. In doing so, we can deduce the causes of infrequent occurrences and determine which actions by sub-systems, robots, or simulators triggered them. By comprehending causality in this manner, we can uncover the underlying reasons behind rare events and plan mitigation procedures accordingly.

\item Process knowledge workflows for rare event prediction

Another direction for rare-event prediction entails the integration of process knowledge workflows into the prediction of rare events. This methodology enhances rare event prediction by integrating domain-specific procedural knowledge with data-driven approaches. This integration has the potential to reveal nuanced abnormalities and disparities that may not be readily apparent, hence leading to more accurate and efficient rare-event prediction.

\item Automated planning for mitigation strategies

Lastly, the road map toward advanced rare-event prediction applications envisions the integration of automated planning techniques. Once rare events are detected and understood, the next step involves devising effective mitigation strategies. Automated planning aims to generate dynamic response plans that can mitigate the impact of rare events. These strategies can encompass various scenarios, offering a comprehensive and adaptable framework for handling unexpected, uncertain occurrences.

\end{enumerate}

In essence, a forward-thinking approach in predicting rare events relies on three crucial components: exploring causes, following knowledge workflows, and implementing automated planning. For instance, predicting rare events in manufacturing assembly pipelines, such as unexpected machinery failures or missing parts, is significant for reducing time and labor costs and improving work processes. Predicting such events involves understanding causality to identify root causes, using process knowledge workflows for accurate prediction, and implementing automated planning strategies to mitigate their impact effectively.

A key insight that we foresee here is that, these components are intricately connected, with each one capable of informing and enhancing the others. For instance, investigating causes can help detect abnormal occurrences and irregularities in knowledge workflows, while insights gained from knowledge workflows can improve automated planning techniques to prevent rare events. By identifying the causes of rare events through exploration, mitigation strategies can be developed in automated planning, and optimized strategies can be used to manipulate variables and gain a better understanding of the causal relationships between factors and rare events. Ultimately, this collaborative approach generates a more comprehensive and holistic framework for predicting rare events across various domains, as information and insights are exchanged between the three elements.

\section{Conclusion}
This paper presents a detailed analysis of rare event prediction, encompassing rare event data, data processing, algorithmic and evaluation approaches. Through our examination of the current literature, we identified several gaps and challenges in the field, highlighting the need for specialized evaluation techniques and their integration. As rare events continue to play a vital role in diverse domains, like manufacturing, healthcare, finance, and earth sciences, addressing these challenges will foster the development of more robust and accurate prediction models. By embracing emerging research trends and leveraging advanced learning methods, we can unlock new opportunities to enhance the prediction and management of rare events, ultimately contributing to safer and more efficient decision-making processes. As the field evolves, interdisciplinary collaboration and innovative solutions will pave the way for transformative advancements in rare event prediction. 

\section{Acronyms}
Table \ref{tab:acronyms} under the Appendix \ref{appendix:F} includes the acronyms used in the paper.

\section{Acknowledgments}
This work is supported in part by NSF grants \#2133842, "EAGER: Advancing Neuro-symbolic AI with Deep Knowledge Infused Learning", and \#2119654, "RII Track 2 FEC: Enabling Factory to Factory (F2F) Networking for Future Manufacturing". Any opinions, findings, conclusions, or recommendations expressed in this material are those of the authors and do not necessarily reflect the views of the NSF.

\bibliographystyle{unsrt}  
\bibliography{references}

\newpage
\appendix
\section*{Appendices}

\section{ACRONYMS: Table \ref{tab:acronyms}}
\label{appendix:F}

\begin{table}[h]
\caption{List of acronyms}
\label{tab:acronyms}
\scriptsize
\centering
\begin{tabular}{>{\hspace{0pt}}m{0.083\linewidth}>{\hspace{0pt}}m{0.367\linewidth}||>{\hspace{0pt}}m{0.092\linewidth}>{\hspace{0pt}}m{0.394\linewidth}} 
\toprule
Acronym & Description & Acronym & Description \\ 
\midrule
N & Numeric data & LIDAR & Light detection and ranging \\
TX & Textual data & LASA & Look-alike-sound-alike \\
I & Image data & FFT & Fast fourier transform \\
A & Audio data & GAN & Generative adversarial networks \\
T & Time series & CGAN & Conditional generative adversarial networks \\
CF & Classification & WGAN & Wasserstein agenerative adversarial networks \\
CL & Clustering & MPM & Mineral prospectivity mapping \\
FT & Forecasting & PCA & Principal component analysis \\
RG & Regression & LR & Logistic regression \\
SM & Simulation & NB & Naive bayes \\
RE & Naturally rare event datasets & NN & Neural networks \\
DE & Derived datasets & CNN & Convolutional neural networks \\
SI & Simulated datasets & MLP & Multi-layer perceptron \\
SY & Synthetic datasets & LSTM & Long short-term memory \\
UCI & University of California Irvine & RF & Random forest \\
KEEL & Knowledge extraction based on evolutionary learning & k-NN & K-nearest neighbors \\
API & Application programming interface & RIPPER & Repeated incremental pruning to produce error reduction \\
DC & Data cleaning & SVM & Support vector machines \\
FS & Feature selection & L.SVM & Support vector machine with linear kernel \\
SL & Sampling & R.SVM & Support vector machine with radial kernel \\
FE & Feature engineering & GSVM & Granular support vector machines \\
ML & Machine learning & RE-WKLR & Rare event weighted kernel logistic regression \\
R1 & Extremely-rare category & MSB & Maximum specificity bias \\
R2 & Very-rare category & IBL & Instance-based learning \\
R3 & Moderately-rare category & 1-NN & 1-Nearest neighbor \\
R4 & Frequently-rare category & PAM & Partition around medoids \\
CoR & Curse of rarity & CLARA & Clustering large applications \\
SFA & Signal fragment assembler & COG & Classification using lOcal clusterinG \\
VAE & Variational autoencoder & BIRCH & Balanced iterative reducing and clustering~ \\
DP & Data picker & \multicolumn{2}{>{\hspace{0pt}}m{0.486\linewidth}}{~using hierarchies} \\
QC & Quality classifier & ARIMA & Autoregressive integrated moving average \\
MOA & Massive online analysis & VAR & Vector autoregression \\
WEKA & Waikato environment for knowledge analysis & GRU & Gated recurrent unit \\
PHQ & Patient health questionnaire & EMM & Extensible markov models \\
HIV & Human immunodeficiency virus & DNO & Deep neural operators \\
WWD & Wrong-way driving & FNACC & Faulty-normal accuracy \\
APS & Air pressure system & RFNACC & Real faulty-normal accuracy \\
SVD & Singular value decomposition & TP & True positives \\
MICE & Multiple imputation by chained equation & FN & False negatives \\
ANOVA & Analysis of variance & TN & True negatives \\
TL & Tomek links & FP & False positives \\
ENN & Edited nearest neighbors & TNR & True negative rate \\
RFE & Recursive feature elimination & FPR & False positive rate \\
HMM & Hidden markov model & FNR & False negative rate \\
DWT & Discrete wavelet transform & G-Mean & Geometric Mean \\
mRMR & Minimum redundancy maximum relevance & BER & Balanced error rate \\
TF-IDF & Term frequency-inverse document frequency & MCC & Matthews’s correlation coefficient \\
XGBoost & eXtreme Gradient Boosting & PR & Precision-Recall \\
MFCC & Mel-frequency cepstral coefficients & AUPRC & Area under precision-recall curve \\
MDI & Mean decrease in impurity & TDL & Top-decile lift \\
ROS & Random minority oversampling & TSS & True skill statistic \\
RUS & Random majority undersampling & SSE & Sum of squared errors \\
SMOTE & Synthetic minority oversampling technique & MAE & Mean absolute error \\
ADASYN & Adaptive synthetic sampling technique & RMSE & Root mean squared error \\
SMUTE & Similarity majority under-sampling technique & MAPE & Mean absolute percentage error \\
NCL & Neighborhood cleaning rule & MAD & Mean absolute deviation \\
NM & NearMiss & MAPD & Mean absolute percentage deviation \\
NM2 & NearMiss-2 & AUC-ROC & Area under the receiver operating~ \\
OSS & One-sided selection & \multicolumn{2}{>{\hspace{0pt}}m{0.486\linewidth}}{~characteristic curve} \\
CBO & Cluster-based oversampling & UQ & Uncertainty quantification \\
 &  & F2F & Factory to Factory \\
\bottomrule
\end{tabular}
\end{table}

\end{document}